%% file: main.tex
\documentclass[10pt,twocolumn,letterpaper]{article}
\usepackage{cvpr}  
\input{preamble}
\usepackage{graphicx}
\usepackage{subcaption}
\usepackage{amsfonts} 
\usepackage{booktabs,siunitx,makecell,subcaption,caption,adjustbox,xcolor,colortbl}
\renewcommand{\arraystretch}{1.12}
\usepackage{tikz}
\usepackage{amsmath}
\usepackage{algorithm}
\usepackage{algpseudocode}
\usepackage{booktabs} 
\usepackage{multirow} 
\usepackage{siunitx}  
\usepackage{stfloats}     
\usepackage{float}        
\usepackage{xcolor}
\usepackage{balance}
\definecolor{cvprblue}{rgb}{0.21,0.49,0.74}
\usepackage[pagebackref,breaklinks,colorlinks,allcolors=cvprblue]{hyperref}
\usepackage{mathtools}
 
\def\paperID{5479} 
\def\confName{CVPR}
\def\confYear{2026}

\definecolor{color1}{rgb}{0.9, 0.95, 1.0}
\cellcolor{color1} 

\makeatletter
\DeclareRobustCommand\onedot{\futurelet\@let@token\@onedot}
\def\@onedot{\ifx\@let@token.\else.\null\fi\xspace}

\def\eg{\emph{e.g}\onedot}

\makeatother

\definecolor{beaublue}{rgb}{0.9, 0.95, 1.0}
\definecolor{blackish}{rgb}{0.2, 0.2, 0.2}

\usepackage{tcolorbox}

\title{POLARIS: Projection-Orthogonal Least Squares for Robust and Adaptive Inversion in Diffusion Models}

\author{
Wenshuo Chen$^{1}$\thanks{Equal contribution.} \quad
Haosen Li$^{1*}$ \quad
Shaofeng Liang$^{1*}$ \quad
Lei Wang$^{2,3}$ \quad
Haozhe Jia$^{1}$ \quad
Kaishen Yuan$^{1}$ \\
Jieming Wu$^{1}$ \quad
Bowen Tian$^{1}$ \quad
Yutao Yue$^{1}$\thanks{Corresponding author (yutaoyue@hkust-gz.edu.cn).} \\
$^{1}$ The Hong Kong University of Science and Technology (Guangzhou) \\
$^{2}$ Griffith University 
$^{3}$ Data61/CSIRO\\
Project: \href{https://polaris-code-official.github.io/}{polaris-code-official.github.io}
}

\let\oldtwocolumn\twocolumn
\renewcommand\twocolumn[1][]{%
    \oldtwocolumn[{#1}{
    \begin{center}
    \vskip-6ex
        \centering
        \includegraphics[width=1.0\textwidth]{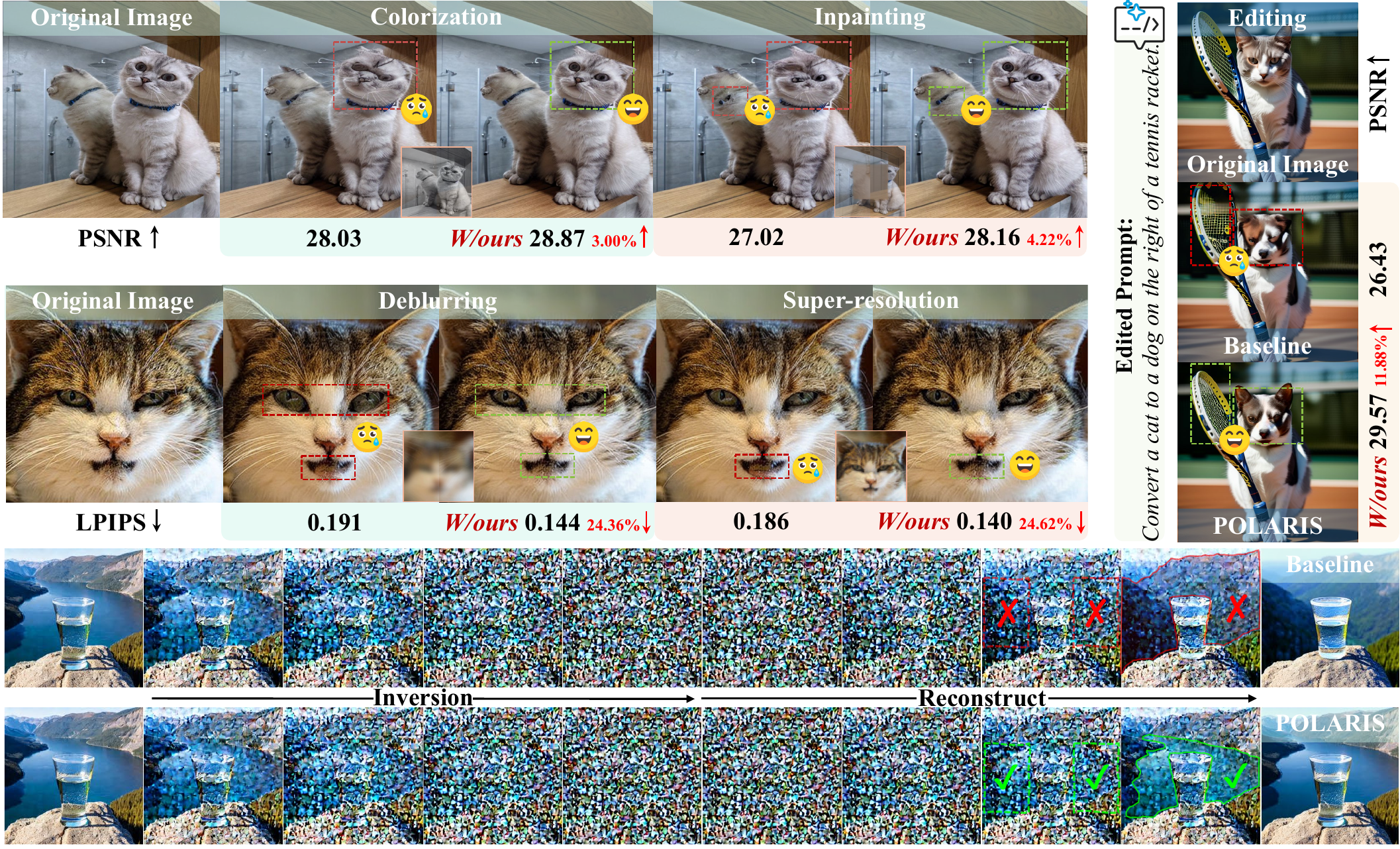}
        \captionsetup{hypcap=false}
        \vskip-2ex
        \captionof{figure} {POLARIS serves as a superior inversion foundation, significantly enhancing the performance of various downstream diffusion tasks. (\textbf{Top two rows}) Image restoration: In colorization, inpainting, deblurring, and super-resolution tasks, all results show significant fidelity improvements when POLARIS is used as the inversion foundation. (\textbf{Right panel}) Image editing: In the ``cat-to-dog" editing task, POLARIS enables the model to successfully execute this complex object replacement. (\textbf{Bottom row}) Reconstruction: POLARIS achieves near-perfect inversion and reconstruction, providing a robust and high-fidelity foundation for downstream tasks.
        }
        \label{fig:1}
    \end{center}
    }]
}

\begin{document}
\maketitle
\let\twocolumn\oldtwocolumn

\input{sec/0_abstract}    
\input{sec/1_introduction}

\input{sec/2_preliminaries}
\input{sec/3_methodology}
\input{sec/4.Experimental_results}
\input{sec/5_Conclusion}

\balance
{
    \small
    \bibliographystyle{unsrt}
    \bibliography{reference}
}

\input{sec/X_suppl}

\end{document}

%% file: preamble.tex
\usepackage{amsthm}  



%% file: sec/0_abstract.tex
\begin{abstract}
The Inversion-Denoising Paradigm, which is based on diffusion models, excels in diverse image editing and restoration tasks. We revisit its mechanism and reveal a critical, overlooked factor in reconstruction degradation: the approximate noise error. This error stems from approximating the noise at step $t$ with the prediction at step $t\!-\!1$, resulting in severe error accumulation throughout the inversion process. We introduce Projection-Orthogonal Least Squares for Robust and Adaptive Inversion (POLARIS), which reformulates inversion from an error-compensation problem into an error-origin problem. Rather than optimizing embeddings or latent codes to offset accumulated drift, POLARIS treats the guidance scale $\omega$ as a step-wise variable and derives a mathematically grounded formula to minimize inversion error at each step. Remarkably, POLARIS improves inversion latent quality with just one line of code. With negligible performance overhead, it substantially mitigates noise approximation errors and consistently improves the accuracy of downstream tasks.

\end{abstract}

%% file: sec/1_introduction.tex
\section{Introduction}

Recent advances in diffusion models 
\cite{ho2020denoisingdiffusionprobabilisticmodels, song2022denoisingdiffusionimplicitmodels, 
rombach2022highresolutionimagesynthesislatent, song2021scorebasedgenerativemodelingstochastic, 
lipman2023flowmatchinggenerativemodeling, jia2025physicsinformedrepresentationalignmentsparse, 
chen2025freet2mfrequencyenhancedtexttomotion, chen_2024, Chen_2025, 
ning2025dctdiffintriguingpropertiesimage} 
have enabled powerful real-world image editing capabilities 
\cite{hertz2022prompt, cao_2023_masactrl, parmar2023zeroshotimagetoimagetranslation, 
tumanyan2022plugandplaydiffusionfeaturestextdriven, Gomez_Trenado_2025, kawar2023imagic, 
meng2022sdedit, couairon2022diffeditdiffusionbasedsemanticimage, yang2022paint, 
Avrahami_2023, brooks2022instructpix2pix}. 
These methods rely on denoising inversion, which has become the core mechanism behind modern 
image manipulation and restoration. 
The dominant methodology for this task hinges on a critical preliminary step: inverting a 
source image into the model's latent space 
\cite{kingma2022autoencodingvariationalbayes, rombach2022highresolutionimagesynthesislatent}. 
This process is commonly realized through deterministic schedulers, such as DDIM 
\cite{song2022denoisingdiffusionimplicitmodels}, which generate a reconstructable latent 
trajectory that underpins subsequent guidance-based editing. 
The success of countless editing applications today is built upon the presumed reliability of 
this inversion process.

However, the very foundation of DDIM inversion rests on a critical approximation. To reverse the diffusion process, one must estimate the noise from a previous step. This slight yet crucial discrepancy between the noise predictions at consecutive timesteps introduces a persistent error that subtly corrupts the latent trajectory. The issue is dramatically amplified by Classifier-Free Guidance (CFG) \cite{ho2022classifierfreediffusionguidance}, an indispensable tool for high-fidelity generation. Practitioners face a difficult dilemma: a low CFG scale struggles to preserve the image's core semantics, while a high scale magnifies the inversion error \cite{mokady2022null, dong2023prompttuninginversiontextdriven}, leading to conspicuous artifacts and a visible departure from the source image \cite{cfg++}.

In response to this challenge, a family of methods \cite{cho2024noisemapguidanceinversion, dong2023prompttuninginversiontextdriven, wallace2022edictexactdiffusioninversion,ju2023directinversionboostingdiffusionbased,miyake2024negativepromptinversionfastimage,mokady2022null,wang2024belm,samuel2025lightningfastimageinversionediting} has emerged. These approaches tacitly accept this foundational flaw in the inversion process and instead focus on designing intricate algorithms to mitigate its downstream effects during the denoising and editing phase. While effective, they often build intricate corrective layers on top of a flawed base, frequently at the cost of implementation complexity, computational efficiency, and mathematical elegance. Furthermore, their corrective nature often leads to suboptimal preservation of the image background during editing, as the model struggles to disentangle editing effects from inversion error compensation.

In this paper, we argue that a more fundamental approach is necessary. Rather than compensating for the consequences of inversion error \cite{mokady2022null,ju2023directinversionboostingdiffusionbased,dong2023prompttuninginversiontextdriven,miyake2024negativepromptinversionfastimage}, we are \textit{the first} to confront its root cause: the inherent discrepancy in noise predictions between consecutive steps. We reframe the challenge, proposing to treat the inversion process as a direct optimization problem. We formally define the inversion error as the Euclidean norm of the noise prediction difference and seek to minimize it. A analytical solution for the optimal CFG scale proves to be fragile and impractical. Our key insight, however, is that by strategically pruning negligible historical dependency terms from the objective, we can derive \textit{a rigorous and robust optimization}. The solution yields an adaptive, optimal CFG scale for each inversion step, a method we call Projection-Orthogonal Least Squares for Robust and Adaptive Inversion (POLARIS). This approach is remarkably simple, effective, and intuitive, achieving near-perfect inversion with a dynamic CFG scale adjustment implemented in just one line of code.


We demonstrate the power and versatility of POLARIS through extensive experiments on massive benchmarks such as
Pick-a-Pic~\cite{kirstain2023pickapicopendatasetuser} and COCO 2017~\cite{lin2015microsoftcococommonobjects}, covering
diverse editing and restoration tasks. Integrated into multiple models, POLARIS serves as a seamless,
plug-and-play module that substantially improves reconstruction fidelity and editing quality with negligible overhead.


Our \textbf{main contributions} are:
\renewcommand{\labelenumi}{\roman{enumi}.}
\begin{enumerate}[leftmargin=0.6cm]
\item We identify a fundamental yet overlooked source of error in denoising inversion: the noise-prediction discrepancy between consecutive steps. This is the root cause of trajectory drift amplified by fixed CFG.
\item We propose POLARIS, which reformulates inversion from error compensation to error-origin optimization by treating the guidance scale as a dynamic per step variable and deriving a closed form update rule that minimizes inversion error at its source.
\item Extensive quantitative and qualitative experiments confirm that POLARIS integrates seamlessly into existing diffusion pipelines, consistently achieving nearly state-of-the-art inversion performance with virtually negligible additional computational cost.
\end{enumerate}

%% file: sec/2_preliminaries.tex
\section{Related Work}
\label{sec:related}

\noindent\textbf{Diffusion model inversion.}
Diffusion models are the standard backbone for high-fidelity image synthesis, and \emph{inversion} is the bridge to editable generative trajectories.
Deterministic DDIM inversion \cite{song2022denoisingdiffusionimplicitmodels} aims to reverse the sampling ODE \cite{song2021scorebasedgenerativemodelingstochastic, wang2024belm}, but coupling with CFG breaks its near-bijection.
Specifically, the fixed guidance scale ($\omega > 1$) amplifies prediction errors inherent in the guidance vector (the difference between conditional and unconditional estimates), causing a compounding trajectory drift.
To compensate, optimization-based schemes like \emph{Null-Text Inversion} \cite{mokady2022null} and \emph{Negative Prompt Inversion} \cite{miyake2024negativepromptinversionfastimage} adjust the conditioning, while others optimize the latent code \cite{wallace2022edictexactdiffusioninversion}.

Despite progress, these methods address the \emph{consequences} of inversion error (via per-sample optimization) rather than its \emph{origin} in the guidance-coupled reverse map.

\noindent\textbf{Editing and restoration via inversion.}
A faithful inverted latent unlocks a broad family of downstream tasks.
For editing, \emph{Prompt to Prompt} \cite{hertz2022prompt} manipulates cross-attention to localize semantic changes while preserving layout; instruction-tuned approaches like \emph{InstructPix2Pix} \cite{brooks2022instructpix2pix} enable intuitive text edits; \emph{Plug-and-Play} \cite{tumanyan2022plugandplaydiffusionfeaturestextdriven} and \emph{MasaCtrl} \cite{cao_2023_masactrl} inject features or control attention to enforce structural consistency.
For restoration, inversion acts as a physics-aware prior: \emph{DDRM} \cite{kawar2022denoising} propagates measurement-space corrections, \emph{DDNM} \cite{wang2022zero} decomposes updates into range/null spaces to strictly enforce known constraints, and \emph{Diffusion Posterior Sampling (DPS)} \cite{chung2024diffusionposteriorsamplinggeneral} applies consistency-gradient guidance.

Nevertheless, these pipelines typically rely on fixed or heuristically tuned guidance during inversion, leaving error accumulation unaddressed at its source.

\noindent\textbf{Position of our work.} POLARIS reframes inversion from an \emph{error-compensation} problem to an \emph{error-origin} problem.
Instead of optimizing embeddings or latents to counteract drift, we treat the guidance scale as a step-wise variable and derive a mathematically grounded rule that minimizes inversion error at each step.
Concretely, we obtain a closed-form, per-timestep scale from an orthogonality condition between conditional/unconditional noise-change vectors, yielding a robust, plug-and-play dynamic schedule that can be dropped into standard DDIM inversion.
POLARIS is thus \textit{complementary to prior pipelines}: it preserves their modularity while replacing a brittle, fixed hyperparameter with an adaptive, theoretically justified control law.

\section{Preliminary}
\textbf{Denoising diffusion implicit models.} DDIM provide a determinate sampling process. The denoising step maps a sample $\mathbf{x}_t$ to a cleaner sample $\mathbf{x}_{t-1}$ using a U-Net prediction $\varepsilon(\mathbf{x}_t)$, governed by a schedule $\bar{\alpha}_t$:
\begin{align}
\mathbf{x}_{t\!-\!1} & = \mathfrak{D}(\mathbf{x}_t) \nonumber 
 = \sqrt{1-\bar{\alpha}_{t\!-\!1}}\,\varepsilon(\mathbf{x}_t) \\
 & \!+\! \sqrt{\bar{\alpha}_{t\!-\!1}}\!\left(\frac{\mathbf{x}_t-\sqrt{1-\bar{\alpha}_t}\,\varepsilon(\mathbf{x}_t)}{\sqrt{\bar{\alpha}_t}}\right) 
\end{align}

The corresponding inversion process maps a data sample back to noise. We denote the states along this inversion path with a tilde, \eg, $\tilde{\mathbf{x}}_{t-1}$ and $\tilde{\mathbf{x}}_t$. The process requires an approximation: since the ideal noise prediction $\varepsilon(\tilde{\mathbf{x}}_t)$ cannot be computed directly \cite{song2022denoisingdiffusionimplicitmodels} (as it depends on the unknown $\tilde{\mathbf{x}}_t$), it is approximated by the prediction from the current state, $\varepsilon(\tilde{\mathbf{x}}_{t-1})$. This leads to the practical inversion formula:

\begin{align}
\tilde{\mathbf{x}}_t & = \mathfrak{I}(\mathbf{x}_{t-1}) \nonumber 
 = \frac{\sqrt{\bar{\alpha}_t}}{\sqrt{\bar{\alpha}_{t-1}}}\tilde{\mathbf{x}}_{t-1} \nonumber \\
& \quad + \left(\sqrt{1-\bar{\alpha}_t} - \frac{\sqrt{\bar{\alpha}_t(1-\bar{\alpha}_{t-1})}}{\sqrt{\bar{\alpha}_{t-1}}}\right) \varepsilon(\tilde{\mathbf{x}}_{t-1})
\end{align}

This approximation makes the inversion computationally tractable but also renders the process nearly, rather than perfectly, invertible.

\begin{algorithm}[h!]
\caption{DDIM Inversion-Sampling with POLARIS}
\label{alg:polaris_editing_tabular}
\begin{tabular}{l} 

\textbf{Part 1: Inversion} \\
\hline 
\textbf{Input:} image $\mathbf{I}_0$, source prompt $c_s$, time step $T$ \\
\textbf{Output:} inverted latent $\mathbf{x}_T$,  scale list $\{\omega_t\}_{t=1}^{T-1}$\\
\hline 

\begin{tabular}{@{}l@{\hspace{0.5em}}l@{\hspace{0.5em}}p{0.8\linewidth}} 
1 & &  $\mathbf{x}_0 \leftarrow Encode(\mathbf{I}_0)$ \\
2 & & \textbf{for} $t = 0, \dots, T-1$ \textbf{do} \\
3 & \multicolumn{1}{c|}{} & \hspace{0.5em} \textbf{if} $t = 0$ \textbf{then} \\
4 & \multicolumn{1}{c|}{} & \hspace{1.5em} $\omega_t \leftarrow \omega_0$ \\
5 & \multicolumn{1}{c|}{} & \hspace{0.5em} \textbf{else} \\
6 & \multicolumn{1}{c|}{} & \hspace{1.5em} Compute $\Delta\varepsilon_\varphi, \Delta\varepsilon_{c_s}$ from step $t$ and $t-1$. \\
7 & \multicolumn{1}{c|}{} & \hspace{1.5em} $\textcolor{blue}{\displaystyle \omega_t \leftarrow \frac{\|\Delta\varepsilon_\varphi\|^2 - \Delta\varepsilon_\varphi \cdot  \Delta\varepsilon_{c_s}}{\|\Delta\varepsilon_\varphi -  \Delta\varepsilon_{c_s}\|^2 + \epsilon}}$ \\
8 & \multicolumn{1}{c|}{} & \hspace{1.5em} Record $\omega_t$ in the scale list. \\
9 & \multicolumn{1}{c|}{} & \hspace{0.5em} \textbf{end if} \\
10 & \multicolumn{1}{c|}{} & \hspace{0.5em} 	 $\varepsilon(\tilde{\mathbf{x}}_t)\leftarrow(1-\omega_t)\varepsilon_\varphi(\tilde{\mathbf{x}}_t) + \omega_t\varepsilon_{c_s}(\tilde{\mathbf{x}}_t)$\\
11 & \multicolumn{1}{c|}{} & \hspace{0.5em} $\tilde{\mathbf{x}}_{t+1}\leftarrow\mathfrak{I} (\tilde{\mathbf{x}}_{t})$\\
12& & \textbf{end for} \\
\end{tabular} \\ 

\hline
\textbf{Part 2: Sampling}\\
\hline
\textbf{Input:} latent $\mathbf{x}_T$, target prompt $c_t$, scale list $\{\omega_t\}_{t=1}^{T-1}$ \\
\textbf{Output:} reconstructed image $\mathbf{I'_0}$\\
\hline
\begin{tabular}{@{}l@{\hspace{0.5em}}l@{\hspace{0.5em}}p{0.8\linewidth}}
1 & & \textbf{for} $t = T, \dots, 1$ \textbf{do} \\
2 & \multicolumn{1}{c|}{} & \hspace{0.5em}  $i \leftarrow T - t + 1$\\
2 & \multicolumn{1}{c|}{} & \hspace{0.5em}$\varepsilon(\mathbf{x}_t) \leftarrow (1 - \omega_i)\varepsilon_{\varphi}(\mathbf{x}_t) + \omega_i\varepsilon_{c_t}(\mathbf{x}_t)$\\
3 & \multicolumn{1}{c|}{} & \hspace{0.5em} $\mathbf{x}_{t-1} \leftarrow \mathfrak{D}(\mathbf{x}_t)$ \\
4& & \textbf{end for} \\
5 & & $\mathbf{I'_0} \leftarrow Decode(\mathbf{x}_0)$ \\
\end{tabular} \\
\end{tabular} 
\end{algorithm}

\textbf{Classifier-free guidance.} 
To control the influence of the conditional prompt $c$ on the generation process, we employ CFG \cite{ho2022classifierfreediffusionguidance}. This technique utilizes a model $\varepsilon$ trained to predict noise for both conditional and unconditional inputs. This is typically achieved by randomly replacing the condition $c$ with a null token $\emptyset$ during training.

\begin{figure*}[tbp] 
    \centering
    \includegraphics[width=\textwidth]{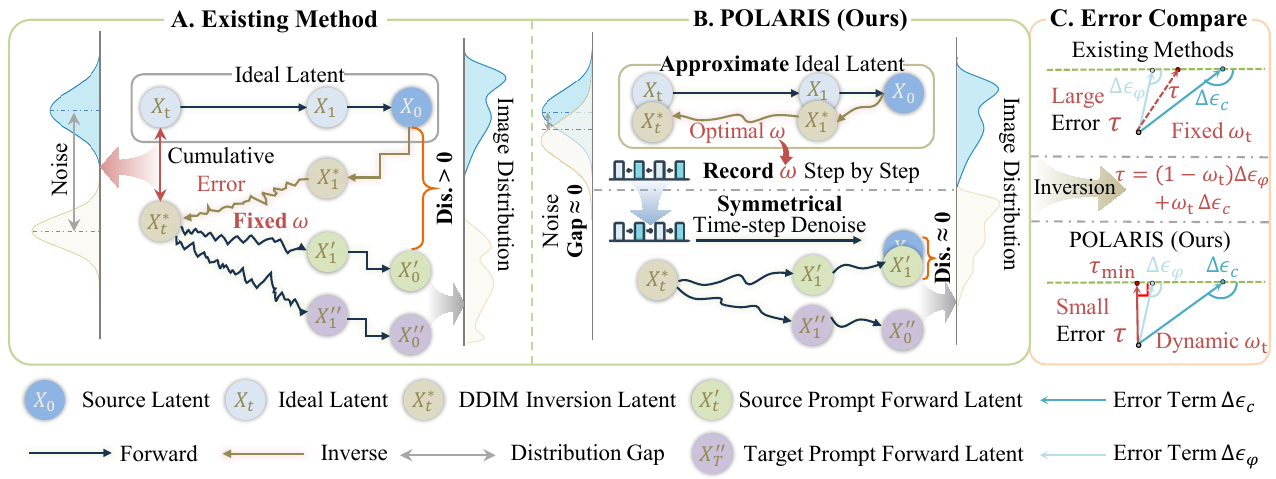}
    \caption{(\textbf{A}) Existing CFG-based DDIM methods introduce and accumulate errors at each step of the inversion process, eventually leading to the distribution shift of downstream tasks. (\textbf{B}) Our method actively seeks a mathematically invertible path to obtain a latent variable that is closer to the ideal one. During generation, the recorded $\omega_t$ sequence is replayed, enabling high-fidelity reconstruction.  (\textbf{C}) Shows our optimization goal from a geometric point of view to minimize the inversion error.} %
    \label{fig:main}
\end{figure*}


At each denoising step $t$, we compute two noise predictions: the unconditional prediction $\varepsilon_\varphi(\mathbf{x}_t)$ and the conditional prediction $\varepsilon_c(\mathbf{x}_t)$. The final guided noise prediction for the denoising process, $\epsilon(\mathbf{x}_t)$, is a linear combination of these two, controlled by a guidance scale $\omega$:
\begin{equation} \label{eq:cfg_formula_denoising}
\varepsilon(\mathbf{x}_t) = (1-\omega)\varepsilon_\varphi(\mathbf{x}_t) + \omega\varepsilon_c(\mathbf{x}_t).
\end{equation}

Similarly, for an inversion path state $\tilde{\mathbf{x}}_{t-1}$, the guided noise $\epsilon(\tilde{\mathbf{x}}_{t-1})$ is computed as:
\begin{equation} \label{eq:cfg_formula_inversion}
\varepsilon(\tilde{\mathbf{x}}_{t-1}) = (1-\omega)\varepsilon_\varphi(\tilde{\mathbf{x}}_{t-1}) + \omega\varepsilon_c(\tilde{\mathbf{x}}_{t-1}).
\end{equation}
The guidance scale $\omega$ is a scalar hyperparameter. With $\omega=0$ the process is unconditional; with $\omega>1$ prompt guidance $c$ is amplified. In standard DDIM sampling/inversion, $\omega$ is typically constant across timesteps.

%% file: sec/3_methodology.tex
\newtheorem{theorem}{Theorem}

\section{Methodology}

Our method aims to minimize the DDIM inversion error by dynamically optimizing the guidance scale $\omega$ at each timestep. We first derive a theoretically exact but practically unstable solution. We then analyze its failure modes and propose a robust approximation grounded in both empirical evidence and theoretical analysis. All formula derivations and proofs can be found in Appendix \ref{sec:proof}.

\subsection{The Exact Solution and Its Instability}

The core of DDIM inversion error stems from the approximation 
$\varepsilon(\tilde{\mathbf{x}}_t)\!\approx\!\varepsilon(\tilde{\mathbf{x}}_{t-1})$.
Our goal is to minimize this discrepancy, which we define as 
the inversion error $\tau_{\text{inv}}(t)$:
\begin{equation}
    \tau_\text{inv}(t) = \varepsilon(\tilde{\mathbf{x}}_t) - \varepsilon(\tilde{\mathbf{x}}_{t-1}),
\end{equation}
where the guided noise at steps $t$ and $t-1$ are given by:
\begin{align}
\left\{
\begin{aligned}
	\epsilon(\tilde{\mathbf{x}}_t) &= (1-\omega_t)\varepsilon_\varphi(\tilde{\mathbf{x}}_t) + \omega_t\varepsilon_c(\tilde{\mathbf{x}}_t) \\
	{\epsilon}(\tilde{\mathbf{x}}_{t-1}) &= (1-\omega_{t-1})\varepsilon_\varphi(\tilde{\mathbf{x}}_{t-1}) + \omega_{t-1}\varepsilon_c(\tilde{\mathbf{x}}_{t-1})
\end{aligned}
\right.
\end{align}

\noindent By defining $\Delta\omega = \omega_t - \omega_{t-1}$, we can expand the error term:
\begin{align}
	\tau_\text{inv}(t) &= (1-\omega_t)(\varepsilon_\varphi(\tilde{\mathbf{x}}_t) - \varepsilon_\varphi(\tilde{\mathbf{x}}_{t-1})) \nonumber \\
	&\quad + \omega_t(\varepsilon_c(\tilde{\mathbf{x}}_t) - \varepsilon_c(\tilde{\mathbf{x}}_{t-1})) \nonumber \\
	&\quad + \Delta\omega(\varepsilon_c(\tilde{\mathbf{x}}_{t-1}) - \varepsilon_\varphi(\tilde{\mathbf{x}}_{t-1})) \nonumber \\
	&= \mathbf{a} + \mathbf{b}\Delta\omega, \label{eq:tau_decomposition}
\end{align}
where we define $\mathbf{a} = (1-\omega_t)\Delta\varepsilon_\varphi + \omega_t\Delta\varepsilon_c$ as the error from prediction changes within the current step, and $\mathbf{b} = \varepsilon_c(\tilde{\mathbf{x}}_{t-1}) - \varepsilon_\varphi(\tilde{\mathbf{x}}_{t-1})$ as the guidance direction from the previous state. The deltas are defined as $\Delta\varepsilon_\varphi = \varepsilon_\varphi(\tilde{\mathbf{x}}_t) - \varepsilon_\varphi(\tilde{\mathbf{x}}_{t-1})$ and $\Delta\varepsilon_c = \varepsilon_c(\tilde{\mathbf{x}}_t) - \varepsilon_c(\tilde{\mathbf{x}}_{t-1})$.

Minimizing $\|\mathbf{a} + \mathbf{b}\Delta\omega\|^2$ with respect to $\Delta\omega$ yields the closed-form solution:
\begin{equation}
\Delta\omega = -\frac{\mathbf{a}\cdot\mathbf{b}}{\|\mathbf{b}\|^2}.
\label{eq:exact_delta_omega}
\end{equation}

This provides the optimal guidance scale for the next step as $\omega_t = \omega_{t-1} + \Delta\omega$.

Given an initial guidance scale $\omega_T$ at the final timestep $T$, the optimization process becomes a sequential problem. At each inversion step $t$ (from $0$ up to $T-1$), the value of $\omega_t$ is known from the calculation of the previous step. Consequently, the only variable in the optimization problem at step $t$ is $\Delta\omega_t$. This allows us to compute the optimal change using Eq.~\eqref{eq:exact_delta_omega}, which in turn determines the guidance scale for the subsequent step, $\omega_{t-1}$. 
This establishes a recursive formula for the guidance scale:
\begin{equation} \label{eq:recursive_omega}
    \omega_{t} = \omega_{t-1} + \Delta\omega_t = \omega_{t-1} - \frac{\mathbf{a}_t \cdot \mathbf{b}_t}{\|\mathbf{b}_t\|^2}
\end{equation}
where the terms $\mathbf{a}_t$ and $\mathbf{b}_t$ are computed at step $t$.

By unrolling this recursive relationship, we can express the guidance scale $\omega_t$ at any arbitrary step $t < T$ as a function of the initial value $\omega_0$ and the sum of all optimal changes from step $1$ up to $T$:
\begin{equation} \label{eq:summation_omega}
    \omega_t = \omega_0 + \sum_{k=1}^{t} \Delta\omega_k = \omega_0 - \sum_{k=1}^{t} \frac{\mathbf{a}_k \cdot \mathbf{b}_k}{\|\mathbf{b}_k\|^2}
\end{equation}
Eq.~\eqref{eq:summation_omega} explicitly shows that the optimal guidance scale at any point is the initial scale adjusted by the cumulative sum of all subsequent error-minimizing changes.

We utilize Eq.~\eqref{eq:summation_omega} to compute the optimal guidance scale for each inversion timestep of the input image. These scales are then recorded and used for denoising at the corresponding timesteps during reconstruction. 

\begin{figure}[t]
    \includegraphics[width=\columnwidth]{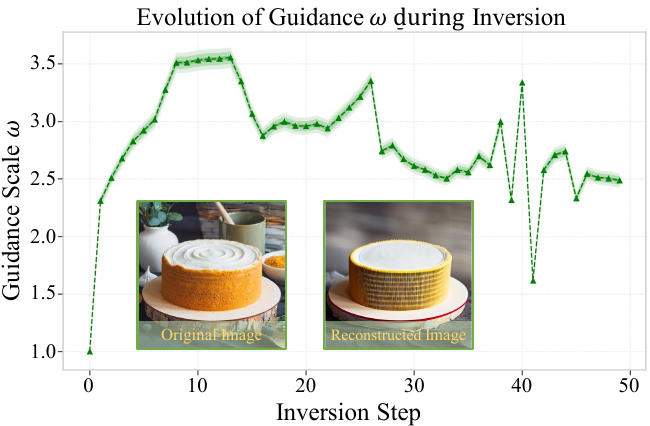}
    \vspace{-0.6cm}
    \caption{Instability of the ``Exact Solution" and resulting reconstruction collapse. The guidance scale $\omega$ computed by this solution (green line) exhibits extreme fluctuations and instability.  It leads to a total collapse in the reconstruction (right), which is filled with artifacts and distortion compared to the original image (left). This phenomenon confirms the solution's ``practical fragility"  and strongly motivates our search for a more robust approximation.}
    \label{fig:exact}
\end{figure}

As shown in Figure~\ref{fig:exact}, the experimental results indicate that our proposed exact dynamic scale method offers no significant improvement in reconstruction quality over the traditional fixed-scale approach; both methods produce images with significant deviations from the original. Moreover, the dynamic scale's evolution curve exhibits extreme fluctuations.
This phenomenon verifies the practical fragility of the exact solution, which often leads to divergent or nonsensical results. \textbf{Its instability stems from a strong dependence on the historical state vector~$\mathbf{b}$}. This dependency can cause errors to accumulate and be catastrophically amplified across time steps. The following theorem, proved in \ref{thm1}, formalizes the instability:
\begin{theorem} \label{prop:exact_instability}
The computation of~$\Delta\omega$ is ill-posed, as small prediction noise $\delta$ can cause its error to diverge when the historical guidance direction~$\|\mathbf{b}\|$ approaches zero.
\end{theorem}

\begin{figure}[t]
    \includegraphics[width=\columnwidth]{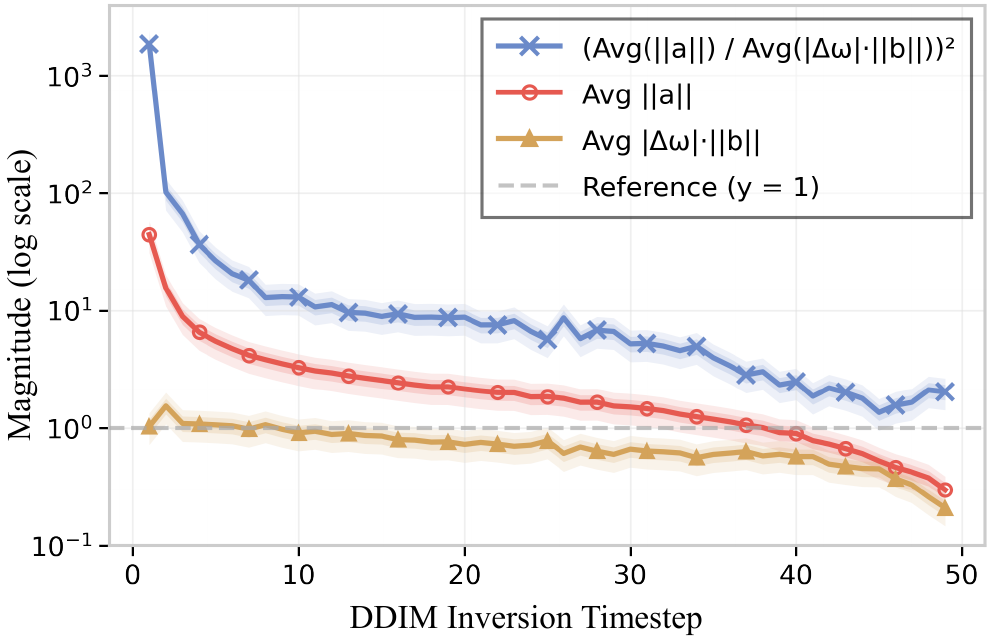}
    \vspace{-0.6cm}
    \caption{Empirical validation of the negligibility of the history-dependent term. It compares the average magnitude of the current-step error term $\|a\|$ (red line) against the history-dependent term $\|\Delta\omega\| \cdot \|b\|$ (yellow line). The squared ratio of their magnitudes (blue line) is shown to be significantly greater than the reference line across the vast majority of timesteps. It validates our core hypothesis that the history term $\mathbf{b}\Delta\omega$ is numerically negligible, justifying our approximation in Eq. (\ref{eq:approx}).}
    \label{fig:magnitude_analysis}
    \vspace{-0.3cm}
\end{figure}

\subsection{A Robust Approximate Solution}

The failure of the exact solution stems from the history-dependent term $\mathbf{b}\Delta\omega$ in Eq.~\eqref{eq:tau_decomposition}. To develop a robust alternative, we hypothesize that this term is numerically negligible compared to the current-step error term $\mathbf{a}$.

To validate this hypothesis, we conducted a batch analysis experiment. 
We randomly selected 1000 samples from the Pick-a-Pic dataset. 
Since our optimization objective is the mean squared error (MSE), 
the contribution of any error component to the total loss is proportional 
to its squared magnitude. 
Therefore, to assess the relative impact of the two error sources, 
at each step $t$ of the DDIM inversion, we computed the ratio of 
squared magnitudes: $(|\mathbf{a}| / |\mathbf{b}\Delta\omega|)^2$.

The results (see Figure~\ref{fig:magnitude_analysis}) provide empirical support, 
showing that the square ratio is significantly greater than 1 across the vast 
majority of steps, averaging over 20. This confirms that $\mathbf{a}$ is the 
dominant component of the inversion error; its average magnitude was 3.2662, 
compared to just 0.7215 for the history-dependent term 
$|\Delta\omega| \cdot \|\mathbf{b}\|$. Hence, the contribution of $\|\mathbf{b}\|$ 
can be considered negligible.

Based on this empirical evidence, we propose an approximation by simplifying 
the optimization to neglect the history-dependent term. 
After that, we define the error:

\begin{equation}
    \tau_\text{approx}(t) = \mathbf{a} = (1-\omega_t)\Delta\varepsilon_\varphi + \omega_t\Delta\varepsilon_c.
    \label{eq:approx}
\end{equation}

Our new goal is to find an optimal $\omega_t$ at each step that independently minimizes $\|\tau_\text{approx}(t)\|^2$.

We derive the solution from a geometric perspective (see Fig.\ref{fig:main}(c)).
The error vector $\tau_{\text{approx}}(t)$ is an affine combination of
$\Delta\varepsilon_\varphi$ and $\Delta\varepsilon_c$, hence its endpoint lies on the line
connecting the endpoints of these two vectors. The minimum-norm point on this line is obtained
by projecting the origin onto it, i.e., by enforcing that $\tau_{\text{approx}}(t)$ is orthogonal
to the line's direction. Let $\vec{v} = \Delta\varepsilon_c - \Delta\varepsilon_\varphi$ denote this direction.
The orthogonality condition reads:
\begin{equation}
    \tau_\text{approx}(t) \cdot (\Delta\varepsilon_c - \Delta\varepsilon_\varphi) = 0.
\end{equation}

\noindent Solving this equation for $\omega_t$ yields proposed robust solution:
\begin{equation} \label{eq:robust_omega_solution}
    \omega(t) = \frac{\|\Delta\varepsilon_\varphi\|^2 - \Delta\varepsilon_\varphi \cdot \Delta\varepsilon_c}{\|\Delta\varepsilon_\varphi - \Delta\varepsilon_c\|^2}.
\end{equation}

This solution is computed at each step $t$ using only information available at that step, thus avoiding the cascading error problem of the exact solution. 

The following theorem which is proved in \ref{thm2} establishes the theoretical robustness of this approximate solution:

\begin{theorem} \label{prop:approx_robustness}
    The computation of $\omega(t)$ is well-posed. Under the mild condition $\|\Delta\varepsilon_\varphi - \Delta\varepsilon_c\| > 0$, its error is bounded and of the same order as the input prediction noise.
\end{theorem}

It is worth noting that the additional computational cost of POLARIS comes entirely from calculating $\omega(t)$ based on known variables. Compared with models containing a large number of parameters, this \textbf{computational overhead is negligible}. See Appendix \ref{sec:efficiency} for details.

%% file: sec/4.Experimental_results.tex
\section{Experiment}

\input{tab/tab1_main}

In this section, we conduct a comprehensive set of experiments to thoroughly evaluate POLARIS. In Sec.~\ref{sec:reconstruction}, we present the most intuitive reconstruction experiments. Sec.~\ref{sec:editing} discusses the performance of POLARIS on the most representative inversion-based tasks. Sec.~\ref{sec:other task} demonstrates the broader applicability of POLARIS across various settings. Finally, Sec.~\ref{sec:Ablation} analyzes whether the scale still admits locally optimal solutions and examines the impact of different initialization strategies.

\subsection{Diffusion Inversion and Reconstruction}
\label{sec:reconstruction}
To validate the effectiveness of POLARIS (Algorithm~\ref{alg:polaris_editing_tabular}), we conducted a large-scale reconstruction evaluation. It serves as the most direct test of inversion quality, since it measures how well a latent code reproduces the original image. We compare POLARIS with a fixed-scale DDIM inversion baseline implemented in Stable Diffusion v1.5 \cite{rombach2022highresolutionimagesynthesislatent}, evaluating both on COCO2017 \cite{lin2015microsoftcococommonobjects} and Pick-a-Pic \cite{kirstain2023pickapicopendatasetuser}. Reconstruction fidelity was measured from 10 to 100 inversion steps. A high guidance scale (set to 7.5)~\cite{mokady2022null} was used during reconstruction to ensure strong conditioning on the source prompt, and we assess its performance using 8-bit MSE, LPIPS \cite{zhang2018unreasonableeffectivenessdeepfeatures}, PSNR \cite{fardo2016formalevaluationpsnrquality}, and SSIM \cite{1284395}.

\begin{figure*}[tbp] 
    
    \centering
    \includegraphics[width=\textwidth]{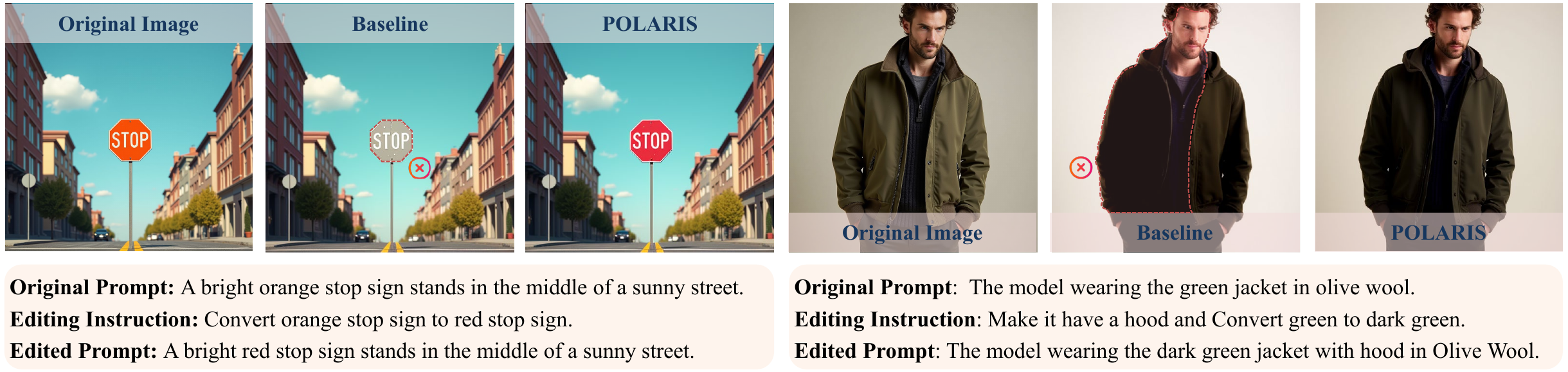}
    \caption{Qualitative comparison of POLARIS on complex image editing tasks.
Compared to the baseline approach, our POLARIS demonstrates superior editing fidelity and the ability to follow complex instructions.}

    \label{fig:six_in_a_row_seamless}
\end{figure*}

As shown in Table~\ref{tab:metric_combined}, POLARIS achieves significant improvements over the baseline on both the COCO2017 and Pick-a-Pic datasets. It indicates that our approach yields a superior latent representation and markedly reduces the error from inversion estimation compared to the baseline. 

The bottom row of Figure~\ref{fig:1} visualizes the 100-step inversion and denoising process. Compared with the baseline, which frequently produces erroneous sampling results, our method yields a better latent noise with richer semantics. Because POLARIS incurs a smaller joint error during denoising and inversion, our model can maintain consistency between the background and the main subject. Moreover, as the number of steps increases, the baseline's cumulative error continues to grow. In contrast, our method exhibits a decreasing error trend, which robustly highlights its ability to mitigate error accumulation.

\input{tab/tab2_editing}

\subsection{Image Editing}
\label{sec:editing}
We evaluate our image editing method on the EditBench dataset~\cite{wang2023imageneditoreditbenchadvancing}, 
focusing on the object-changing sub-task that poses diverse object manipulation challenges.
For all experiments, we use the Stable Diffusion v1.5 model with a DDIM sampler over 50 inference steps.
To provide a comprehensive evaluation, we compare POLARIS from two complementary perspectives: 
\textit{(\textbf{i})} a \emph{pipeline-level} baseline, Prompt-to-Prompt~\cite{hertz2022prompt} (mask), 
as described in Section~\ref{baselines}, to highlight the advancement of our complete inversion-sampling pipeline presented in Algorithm~\ref{alg:polaris_editing_tabular}; and 
\textit{(\textbf{ii})} a \emph{latent-level} baseline, SAGE~\cite{Gomez_Trenado_2025}, which uses only the inverted latent without modifying the editing pipeline, 
to demonstrate the superior inversion quality achieved by POLARIS. 
For fair comparison, the inversion guidance scale of all baseline methods is fixed at $1.0$. 
Quantitatively, we employ the provided masks to measure background preservation, 
and further assess overall visual quality and human preference using the Aesthetic Score.

As shown in Table~\ref{tab:back_metrics_swapped}, the quantitative results demonstrate that POLARIS significantly enhances background preservation across both baseline methods. When integrated with SAGE, our method improves the Back-PSNR by near 1.0 dB and reduces the Back-MSE by nearly 20\%. A similar trend is observed with P2P (Mask), where POLARIS reduces the Back-LPIPS by approximately 40\% and boosts the Back-PSNR to nearly 30 dB. 
This substantial improvement stems from POLARIS's core design: by minimizing the inversion approximation error inherent in baselines, which leads to a much more consistent reconstruction of the background (should remain unchanged).

While our method demonstrates excellent background fidelity, the Aesthetic Score \cite{schuhmann2022laion5bopenlargescaledataset}, a proxy for semantic correctness, does not improve to the same degree. This is intentional, as our approach is engineered to minimize approximation error rather than to directly optimize for semantic accuracy. Therefore, POLARIS is \textit{orthogonal} to methods focused on semantic alignment, making it a complementary module that can be combined with them to boost overall editing performance.

\input{tab/tab3_other_tasks}

\subsection{Extension to Other Inversion Tasks}

\label{sec:other task}

Our experimental setup evaluates four inversion-based downstream tasks: Gaussian deblurring ($\sigma=40.0$), $8\times$ super-resolution, central-mask inpainting, and colorization on Stable Diffusion v1.5 with a DDIM scheduler. All experiments use 50 inference steps at $512\times512$ resolution. To ensure fair comparison, we preserve the original task-specific pipelines; 
the only modification concerns the guidance-scale scheduling. 
In POLARIS, both the inversion and denoising stages employ the dynamic, per-step guidance scale $\omega_t$ 
computed by our method, whereas all baseline methods use a fixed scale of $1.0$ in both stages.

As shown in Table \ref{tab:wide_layout_no_lpips}, POLARIS consistently improves reconstruction fidelity across all inversion-based restoration tasks and baselines. Compared to fixed-scale guidance, POLARIS generally reduces MSE and increases both PSNR and SSIM across most evaluated tasks, demonstrating its effectiveness in mitigating accumulated inversion errors. The consistent gains across DPS, DDRM, and DDNM further indicate that POLARIS generalizes well across architectures and degradation types, validating its plug-and-play applicability to diverse diffusion-based restoration pipelines.

\subsection{Ablation Study}
\label{sec:Ablation}

\textbf{Scale initialization.} 
To investigate the sensitivity of POLARIS to the initial guidance scale, $\omega_0$, we conduct an ablation study on 100 samples randomly selected from the COCO 2017 validation set. For each image, we perform a 50-step DDIM reconstruction with using POLARIS   aligned in Algorithm~\ref{alg:polaris_editing_tabular}. The core of the experiment involves manually setting the initial guidance scale $\omega_0$ for the first step of the inversion process to a range of predefined values: $\{0.0, 1.0, 2.0, 3.0, 4.0, 5.0, 7.5, 9.0, 10.0\}$. For each fixed initial value, all subsequent guidance scales ($\omega_{1}$ through $\omega_{T-1}$) are computed dynamically using the Eq. \eqref{eq:robust_omega_solution}. We then evaluate the final reconstruction fidelity using SSIM and LPIPS to assess the impact of different initializations.

\begin{figure}[tbp]
    \includegraphics[width=\columnwidth]{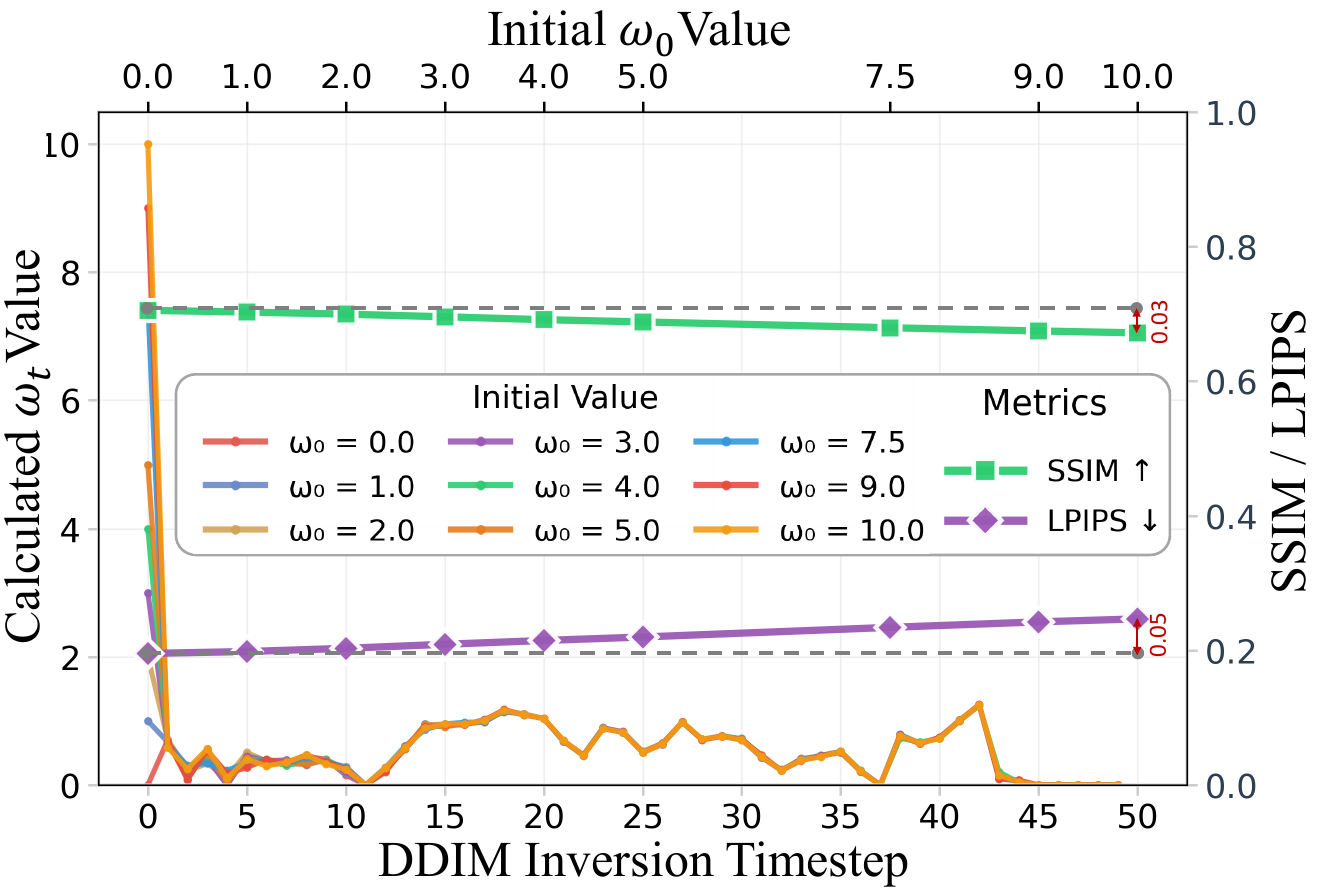}
    \vspace{-0.6cm}
    \caption{Sensitivity Analysis of the Initial Guidance Scale $\omega_0$, demonstrating the robustness of POLARIS.}
    \label{fig:omega}
    \vspace{-0.6cm}
\end{figure}
\noindent\textbf{Optimality verification.}
We observe that the dynamic guidance scales computed by POLARIS typically oscillate within a narrow band around 1.0 between 0 and 2. It raises a critical question: \textbf{Does POLARIS’s performance stem from} its precise, error-minimizing updates at each step, \textbf{or merely from} the generic oscillatory behavior it exhibits?

\begin{figure}[tbp] 
    \centering
    \includegraphics[width=\columnwidth]{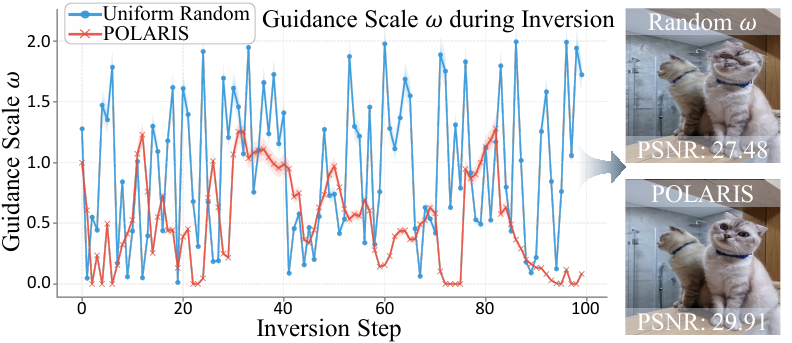}
    \caption{Comparison of the inversion effects of POLARIS and uniform random scale $\omega$. It confirms that the precise, error-minimizing trajectory of POLARIS (red line) is crucial, not just its oscillatory behavior (blue line).}
    \label{fig:compare_2}
\end{figure}

To investigate this, we design a control experiment where, instead of optimization, we sample the guidance scale $\omega_t$ uniformly at random from the interval $U\sim(0, 2)$ at each inversion step $t$. This baseline mimics the oscillatory pattern but lacks the intelligent structure of POLARIS. The entire process is conducted over 50 steps. 
We compare the reconstruction fidelity of POLARIS against this random sampling approach (Figure~\ref{fig:compare_2}). This experiment investigates whether our step-wise optimization truly yields a superior trajectory, or if any similar unstructured oscillation would suffice.
A significant performance gap would confirm that the specific, error-minimizing trajectory found by POLARIS is crucial for high-fidelity reconstruction.

%% file: tab/tab1_main.tex
\begin{table*}[t]
  \centering
  \caption{Performance comparison of POLARIS and Fixed scale across various inference steps on Pick-a-Pic and COCO2017.}
  \label{tab:metric_combined}
  \begin{adjustbox}{max width=\textwidth}
  \begin{tabular}{l ll *{10}{c}}
    \toprule
    \textbf{Dataset} & \textbf{Metric} & \textbf{Method} & \textbf{10} & \textbf{20} & \textbf{30} & \textbf{40} & \textbf{50} & \textbf{60} & \textbf{70} & \textbf{80} & \textbf{90} & \textbf{100} \\
    \midrule

\multirow{8}{*}{\textbf{COCO2017}} 
    & \multirow{2}{*}{MSE ($\downarrow$)} & Fixed scale &
      1827 & 2285 & 2417 & 2614 & 2692 & 2429 & 2617 & 2492 & 2764 & 2866 \\
    & & \cellcolor{color1}POLARIS &
      \cellcolor{color1}506 & \cellcolor{color1}498 & \cellcolor{color1}490 & \cellcolor{color1}493 & \cellcolor{color1}498 &
      \cellcolor{color1}467 & \cellcolor{color1}477 & \cellcolor{color1}464 & \cellcolor{color1}478 & \cellcolor{color1}476 \\
    \addlinespace[0.6ex]
    & \multirow{2}{*}{PSNR  ($\uparrow$)} & Fixed scale &
      15.94 & 14.94 & 14.68 & 14.32 & 14.19 & 14.67 & 14.33 & 14.55 & 14.08 & 13.92 \\
    & & \cellcolor{color1}POLARIS &
      \cellcolor{color1}22.04 & \cellcolor{color1}22.24 & \cellcolor{color1}22.36 & \cellcolor{color1}22.36 & \cellcolor{color1}22.34 &
      \cellcolor{color1}22.62 & \cellcolor{color1}22.54 & \cellcolor{color1}22.66 & \cellcolor{color1}22.53 & \cellcolor{color1}22.55 \\
    \addlinespace[0.6ex]
    & \multirow{2}{*}{SSIM  ($\uparrow$)} & Fixed scale &
      0.5447 & 0.5079 & 0.4959 & 0.4851 & 0.4806 & 0.4869 & 0.4793 & 0.4824 & 0.4723 & 0.4691 \\
    & & \cellcolor{color1}POLARIS &
      \cellcolor{color1}0.6922 & \cellcolor{color1}0.7043 & \cellcolor{color1}0.7085 & \cellcolor{color1}0.7087 & \cellcolor{color1}0.7088 &
      \cellcolor{color1}0.7135 & \cellcolor{color1}0.7122 & \cellcolor{color1}0.7143 & \cellcolor{color1}0.7126 & \cellcolor{color1}0.7133 \\
    \addlinespace[0.6ex]
    & \multirow{2}{*}{LPIPS ($\downarrow$)} & Fixed scale &
      0.4583 & 0.5062 & 0.5181 & 0.5330 & 0.5380 & 0.5206 & 0.5337 & 0.5255 & 0.5440 & 0.5498 \\
    & & \cellcolor{color1}POLARIS &
      \cellcolor{color1}0.2231 & \cellcolor{color1}0.2039 & \cellcolor{color1}0.1957 & \cellcolor{color1}0.1950 & \cellcolor{color1}0.1955 &
      \cellcolor{color1}0.1870 & \cellcolor{color1}0.1897 & \cellcolor{color1}0.1862 & \cellcolor{color1}0.1892 & \cellcolor{color1}0.1886 \\
        \addlinespace[0.8ex]
    \midrule
    \multirow{8}{*}{\textbf{Pick-a-Pic}} 
    & \multirow{2}{*}{MSE ($\downarrow$)} & Fixed scale &
      1688 & 2514 & 2520 & 2530 & 2583 & 2464 & 2537 & 2535 & 2646 & 2691 \\
    & & \cellcolor{color1}POLARIS &
      \cellcolor{color1}632 & \cellcolor{color1}539 & \cellcolor{color1}510 & \cellcolor{color1}527 & \cellcolor{color1}533 &
      \cellcolor{color1}518 & \cellcolor{color1}521 & \cellcolor{color1}519 & \cellcolor{color1}523 & \cellcolor{color1}514 \\
    \addlinespace[0.6ex]
    & \multirow{2}{*}{PSNR ($\uparrow$)} & Fixed scale &
      15.86 & 14.13 & 14.12 & 14.10 & 14.01 & 14.21 & 14.09 & 14.09 & 13.91 & 13.83 \\
    & & \cellcolor{color1}POLARIS &
      \cellcolor{color1}20.12 & \cellcolor{color1}20.81 & \cellcolor{color1}21.05 & \cellcolor{color1}20.91 & \cellcolor{color1}20.87 &
      \cellcolor{color1}20.99 & \cellcolor{color1}20.96 & \cellcolor{color1}20.98 & \cellcolor{color1}20.94 & \cellcolor{color1}21.02 \\
    \addlinespace[0.6ex]
    & \multirow{2}{*}{SSIM ($\uparrow$)} & Fixed scale &
      0.4204 & 0.3442 & 0.3485 & 0.3500 & 0.3485 & 0.3558 & 0.3519 & 0.3506 & 0.3470 & 0.3460 \\
    & & \cellcolor{color1}POLARIS &
      \cellcolor{color1}0.5605 & \cellcolor{color1}0.5992 & \cellcolor{color1}0.6128 & \cellcolor{color1}0.6140 & \cellcolor{color1}0.6154 &
      \cellcolor{color1}0.6204 & \cellcolor{color1}0.6210 & \cellcolor{color1}0.6219 & \cellcolor{color1}0.6215 & \cellcolor{color1}0.6239 \\
    \addlinespace[0.6ex]
    & \multirow{2}{*}{LPIPS ($\downarrow$)} & Fixed scale &
      0.3670 & 0.5041 & 0.4958 & 0.5020 & 0.5056 & 0.5037 & 0.5065 & 0.5192 & 0.5095 & 0.5091 \\
    & & \cellcolor{color1}POLARIS &
      \cellcolor{color1}0.1627 & \cellcolor{color1}0.1213 & \cellcolor{color1}0.1115 & \cellcolor{color1}0.1138 & \cellcolor{color1}0.1138 &
      \cellcolor{color1}0.1101 & \cellcolor{color1}0.1109 & \cellcolor{color1}0.1096 & \cellcolor{color1}0.1123 & \cellcolor{color1}0.1102 \\

    \addlinespace[0.8ex]
    
    \bottomrule
  \end{tabular}
  \end{adjustbox}
\end{table*}

%% file: tab/tab2_editing.tex
\begin{table}[t] 
\centering
\caption{Quantitative comparison of editing on reconstruction, aesthetic, and preference metrics. \textbf{Bold*} indicates the method is enhanced with POLARIS. \underline{Underlined} denotes a mask-based method.}
\label{tab:back_metrics_swapped}
\sisetup{detect-weight,mode=text}

\renewcommand{\arraystretch}{1.0} 
\setlength{\tabcolsep}{3pt} 
\footnotesize 

\begin{tabular}{
  l
  S[table-format=3.3]  
  S[table-format=2.4]  
  S[table-format=1.4]  
  S[table-format=1.4]  
  S[table-format=1.4]  
  c                    
}
\toprule
\textbf{Method} &
\multicolumn{1}{c}{\makecell{\textbf{MSE}\\($\downarrow$)}} &
\multicolumn{1}{c}{\makecell{\textbf{PSNR}\\($\uparrow$)}} &
\multicolumn{1}{c}{\makecell{\textbf{SSIM}\\($\uparrow$)}} &
\multicolumn{1}{c}{\makecell{\textbf{LPIPS}\\($\downarrow$)}} &
\multicolumn{1}{c}{\makecell{\textbf{Aesthetic}\\\textbf{score} ($\uparrow$)}} &
\multicolumn{1}{c}{\makecell{\textbf{Preference}\\\textbf{Rate} (\%)}} \\
\midrule

SAGE    & 266.578 & 23.8728 & 0.8409 & 0.1172 & 5.7392 & 37 \\
\cellcolor{color1}\textbf{SAGE*} & \cellcolor{color1}\bfseries  216.464 & \cellcolor{color1}\bfseries 24.7769 & \cellcolor{color1}\bfseries 0.8725 & \cellcolor{color1}\bfseries 0.1031 & \cellcolor{color1}\bfseries 5.7484 & \cellcolor{color1}\bfseries 63 \\
P2P   & 737.090 & 15.3488 & 0.6057 & 0.3810 & 5.5636 & --- \\
\underline{P2P}   & 79.239 & 29.1414 & 0.9056 & 0.0699 & 5.6842 & 45 \\
\cellcolor{color1}\textbf{\underline{P2P}*} & \cellcolor{color1}\bfseries 65.566 & \cellcolor{color1}\bfseries 29.9640 & \cellcolor{color1}\bfseries 0.9099 & \cellcolor{color1}\bfseries 0.0415 & \cellcolor{color1}\bfseries 5.7315 & \cellcolor{color1}\bfseries 55 \\
\bottomrule
\end{tabular}
\end{table}

%% file: tab/tab3_other_tasks.tex
\begin{table*}[t]
\centering
\caption{Quantitative results comparison across deblurring, super-resolution, colorization, and inpainting. Our method, POLARIS, consistently improves performance over fixed-scale guidance across various tasks and baseline models. The ``---" indicates that the Colorization task is not supported by the original DDNM architecture.}
\label{tab:wide_layout_no_lpips} 

\sisetup{detect-weight, mode=text} 
\renewcommand{\arraystretch}{1.1} 
\setlength{\tabcolsep}{2.5pt}

\newcolumntype{A}{S[table-format=3.2]} 
\newcolumntype{B}{S[table-format=2.2]} 
\newcolumntype{C}{S[table-format=0.3]} 

\begin{tabular}{cc *{4}{A B C}} 
\toprule
\multirow{2}{*}{\textbf{Baseline}} & \multirow{2}{*}{\textbf{Method}}
&  \multicolumn{3}{c}{\textbf{Deblurring}} & \multicolumn{3}{c}{\textbf{Super-resolution}} & \multicolumn{3}{c}{\textbf{Colorization}} & \multicolumn{3}{c}{\textbf{Inpainting}} \\
\cmidrule(lr){3-5} \cmidrule(lr){6-8} \cmidrule(lr){9-11} \cmidrule(lr){12-14} 
 & &
{MSE$\downarrow$} & {PSNR$\uparrow$} & {SSIM$\uparrow$} & 
{MSE$\downarrow$} & {PSNR$\uparrow$} & {SSIM$\uparrow$} & 
{MSE$\downarrow$} & {PSNR$\uparrow$} & {SSIM$\uparrow$} & 
{MSE$\downarrow$} & {PSNR$\uparrow$} & {SSIM$\uparrow$} \\
\midrule

\multirow{2}{*}{DPS \cite{chung2024diffusionposteriorsamplinggeneral}} & Fixed scale & 327.32 & 23.98 & 0.798 & 365.76 & 23.51 & 0.800 & 761.29 & 20.09 & \bfseries 0.719 & 690.69 & 20.45 & 0.719 \\
& \cellcolor{color1}\textbf{POLARIS} & \cellcolor{color1} \bfseries 300.25 & \cellcolor{color1}\bfseries 24.38 &\cellcolor{color1} \bfseries 0.805 &\cellcolor{color1}\bfseries 286.64 & \cellcolor{color1} \bfseries 24.58 &\cellcolor{color1} \bfseries 0.819 &\cellcolor{color1} \bfseries 755.96 & \cellcolor{color1} \bfseries 20.14 &\cellcolor{color1} 0.716 &\cellcolor{color1} \bfseries 632.57 &\cellcolor{color1} \bfseries 20.85 &\cellcolor{color1} \bfseries 0.723 \\
\cmidrule(l){1-14} 

\multirow{2}{*}{DDRM \cite{kawar2022denoising}} & Fixed scale & 284.43 & 24.57 & 0.819 & 329.97 & 24.00 & 0.807 & 703.60 & 20.61 & 0.732 & 566.30 & 21.39 & 0.738 \\
&\cellcolor{color1} \textbf{POLARIS} &\cellcolor{color1} \bfseries 247.31 &\cellcolor{color1} \bfseries 25.18 &\cellcolor{color1} \bfseries 0.828 &\cellcolor{color1} \bfseries 246.24 &\cellcolor{color1} \bfseries 25.33 &\cellcolor{color1} \bfseries 0.830 &\cellcolor{color1} \bfseries 639.17 &\cellcolor{color1} \bfseries 21.00 &\cellcolor{color1} \bfseries 0.735 & \cellcolor{color1}\bfseries 489.51 &\cellcolor{color1} \bfseries 22.06 &\cellcolor{color1} \bfseries 0.746 \\
\cmidrule(l){1-14} 

\multirow{2}{*}{DDNM \cite{wang2022zero}} & Fixed scale & 330.44 & 23.95 & 0.792 & 344.62 & 23.64 & 0.802 & {---} & {---} & {---} & 735.80 & 20.21 & 0.712 \\
&\cellcolor{color1} \textbf{POLARIS} &\cellcolor{color1} \bfseries 320.26 &\cellcolor{color1} \bfseries 24.12 &\cellcolor{color1} \bfseries 0.794 & \cellcolor{color1}\bfseries 290.28 &\cellcolor{color1} \bfseries 24.41 &\cellcolor{color1} \bfseries 0.815 &\cellcolor{color1} {---} &\cellcolor{color1} {---} &\cellcolor{color1} {---} &\cellcolor{color1} \bfseries 700.72 &\cellcolor{color1} \bfseries 20.49 &\cellcolor{color1} \bfseries 0.714 \\
\bottomrule 
\end{tabular}
\end{table*}

%% file: sec/5_Conclusion.tex
\section{Conclusion}
We introduced POLARIS, a novel method that addresses the fundamental problem of approximate noise error in DDIM inversion. By reformulating the process as a per-step optimization and treating the guidance scale as a dynamic variable, POLARIS minimizes error at its source, thereby preventing the error propagation common in standard methods. 
Our experiments show this seamless, plug-and-play approach achieves substantial improvements in reconstruction fidelity and background preservation across editing tasks, with virtually no computational overhead.
While POLARIS excels at object and attribute manipulation, future work will focus on extending its application to complex, non-rigid edits, a known challenge for existing models. Promising future directions also include adapting the POLARIS framework for video and 3D inversion and exploring methods to stabilize its theoretically exact solution to unlock even greater fidelity, further broadening its impact.

%% file: sec/X_suppl.tex
\setcounter{page}{1}
\maketitlesupplementary

\section{Theoretical Analysis}
\label{sec:proof}

\subsection{Theorem 1: Inherent Flaws of the Exact Solution}
\label{thm1}
\begin{proof}
Let the noise prediction at any step be the sum of the true signal and a perturbation term:
\[
\varepsilon_c(\mathbf{x}_t) = \varepsilon_c^*(\mathbf{x}_t) + \delta_c(\mathbf{x}_t).
\]
The main computational terms in the exact solution are thus perturbed as follows:
\[
\mathbf{a} = (1-\omega_t)\Delta\varepsilon_\varphi + \omega_t\Delta\varepsilon_c,\quad
\mathbf{b} = \varepsilon_c(\tilde{\mathbf{x}}_{t-1}) - \varepsilon_\varphi(\tilde{\mathbf{x}}_{t-1}),
\]
\[
\mathbf{a} = \mathbf{a}^* + \delta_a,\quad
\mathbf{b} = \mathbf{b}^* + \delta_b.
\]

	Let
	\[
	\Delta\omega_{\mathrm{obs}} = -\frac{(\mathbf{a}^*+\delta_a)\!\cdot\!(\mathbf{b}^*+\delta_b)}{\|\mathbf{b}^*+\delta_b\|^2},\quad
	\Delta\omega_{\mathrm{true}} = -\frac{\mathbf{a}^*\!\cdot\!\mathbf{b}^*}{\|\mathbf{b}^*\|^2},
	\]
	and the error be \(E(\Delta\omega)=\Delta\omega_{\mathrm{obs}}-\Delta\omega_{\mathrm{true}}\). Then:
	
	\medskip
	\noindent\textbf{(I) Full rational form of the error.} Combining the terms over a common denominator yields
	\begin{equation}\label{eq:full_rational_E}
		E(\Delta\omega)
		=
		\frac{
			(\mathbf{a}^*\!\cdot\!\mathbf{b}^*)\|\delta_b\|^2
			+ 2(\mathbf{a}^*\!\cdot\!\mathbf{b}^*)(\mathbf{b}^*\!\cdot\!\delta_b)
			- (\mathbf{a}^*\!\cdot\!\delta_b)\|\mathbf{b}^*\|^2
			- (\delta_a\!\cdot\!\mathbf{b}^*)\|\mathbf{b}^*\|^2
			- (\delta_a\!\cdot\!\delta_b)\|\mathbf{b}^*\|^2
		}{
			\|\mathbf{b}^*\|^2\,\|\mathbf{b}^*+\delta_b\|^2}.
	\end{equation}
	
	\noindent\textbf{(II) Limit analysis and divergence.} Consider the parameterized path \(\mathbf{b}^*=t\mathbf{u}\) (where \(\|\mathbf{u}\|=1,\,t\to0^+\)), and let \(E(t)=N(t)/D(t)\). Expanding the numerator and denominator from \eqref{eq:full_rational_E} gives
	\[
	N(t)=t(\mathbf{a}^*\!\cdot\!\mathbf{u})\|\delta_b\|^2
	+ 2t^2(\mathbf{a}^*\!\cdot\!\mathbf{u})(\mathbf{u}\!\cdot\!\delta_b)
	- t^2(\mathbf{a}^*\!\cdot\!\delta_b)
	- t^3(\delta_a\!\cdot\!\mathbf{u})
	- t^2(\delta_a\!\cdot\!\delta_b),
	\]
	\[
	D(t)=\|t\mathbf{u}\|^2\,\|t\mathbf{u}+\delta_b\|^2
	=t^4+2t^3(\mathbf{u}\!\cdot\!\delta_b)+t^2\|\delta_b\|^2.
	\]
	
	\noindent As \(t\to0^+\), this is an indeterminate form of \(0/0\). Applying L'H\^opital's rule:
	\[
	N'(t)=(\mathbf{a}^*\!\cdot\!\mathbf{u})\|\delta_b\|^2
	+ 4t(\mathbf{a}^*\!\cdot\!\mathbf{u})(\mathbf{u}\!\cdot\!\delta_b)
	- 2t(\mathbf{a}^*\!\cdot\!\delta_b)
	- 3t^2(\delta_a\!\cdot\!\mathbf{u})
	- 2t(\delta_a\!\cdot\!\delta_b),
	\]
	\[
	D'(t)=4t^3+6t^2(\mathbf{u}\!\cdot\!\delta_b)+2t\|\delta_b\|^2.
	\]
	
	\begin{enumerate}
		\item If \((\mathbf{a}^*\!\cdot\!\mathbf{u})\neq0\), then
		\[
		\lim_{t\to0^+}N'(t)=(\mathbf{a}^*\!\cdot\!\mathbf{u})\|\delta_b\|^2\neq0,\quad
		\lim_{t\to0^+}D'(t)=0,
		\]
		and therefore \(\lim_{t\to0^+}E(t)=\pm\infty\), meaning the error diverges.
		
		\item If \((\mathbf{a}^*\!\cdot\!\mathbf{u})=0\), then \(\lim_{t\to0^+}N'(t)=\lim_{t\to0^+}D'(t)=0\). Applying L'H\^opital's rule a second time:
		\[
		N''(t)=4(\mathbf{a}^*\!\cdot\!\mathbf{u})(\mathbf{u}\!\cdot\!\delta_b)
		- 2(\mathbf{a}^*\!\cdot\!\delta_b)
		- 6t(\delta_a\!\cdot\!\mathbf{u})
		- 2(\delta_a\!\cdot\!\delta_b),
		\]
		\[
		D''(t)=12t^2+12t(\mathbf{u}\!\cdot\!\delta_b)+2\|\delta_b\|^2.
		\]
		Taking the limit under the condition that \((\mathbf{a}^*\!\cdot\!\mathbf{u})=0\) yields
		\[
		\lim_{t\to0^+}E(t)=
		\frac{- 2(\mathbf{a}^*\!\cdot\!\delta_b) - 2(\delta_a\!\cdot\!\delta_b)}{2\|\delta_b\|^2}
		= -\frac{(\mathbf{a}^*\!\cdot\!\delta_b)+(\delta_a\!\cdot\!\delta_b)}{\|\delta_b\|^2}.
		\]
		If we further assume \(\delta_b \rightarrow \mathbf{0}\), then from \eqref{eq:full_rational_E}, we find that \(E(t)\sim-(\delta_a\!\cdot\!\mathbf{u})/t\), which diverges.
	\end{enumerate}
	
	Hence, the exact solution is ill-conditioned as \(\|\mathbf{b}^*\|\!\to\!0\); the error magnitude grows proportionally to \(\|\mathbf{b}^*\|^{-1}\).
\end{proof}

\subsection{Theorem 2: Inherent Advantages of the Approximate Solution}
\label{thm2}
\begin{proof}
	Let the true signals be \(\Delta\varepsilon_\varphi^*, \Delta\varepsilon_c^*\), and the noisy signals be
	\[
	\Delta\varepsilon_\varphi = \Delta\varepsilon_\varphi^* + \Delta\delta_\varphi,\quad
	\Delta\varepsilon_c = \Delta\varepsilon_c^* + \Delta\delta_c,
	\]
	The true and observed values of the approximate solution are defined as
	\[
	\omega_{\mathrm{true}} = \frac{\|\Delta\varepsilon_\varphi^*\|^2 - \Delta\varepsilon_\varphi^* \cdot \Delta\varepsilon_c^*}
	{\|\Delta\varepsilon_\varphi^* - \Delta\varepsilon_c^*\|^2}, \quad
	\omega_{\mathrm{obs}} = \frac{\|\Delta\varepsilon_\varphi\|^2 - \Delta\varepsilon_\varphi \cdot \Delta\varepsilon_c}
	{\|\Delta\varepsilon_\varphi - \Delta\varepsilon_c\|^2}.
	\]
    Let
	\[
	N(\Delta\varepsilon_\varphi, \Delta\varepsilon_c) = \|\Delta\varepsilon_\varphi\|^2 - \Delta\varepsilon_\varphi \cdot \Delta\varepsilon_c, \quad
	D(\Delta\varepsilon_\varphi, \Delta\varepsilon_c) = \|\Delta\varepsilon_\varphi - \Delta\varepsilon_c\|^2.
	\]
	Denote by \(N^*\) and \(D^*\) the values for the true signals, respectively. Then
	\[
	N_{\mathrm{obs}} = N^* + \delta_N + O(\|\Delta\delta\|^2), \quad
	D_{\mathrm{obs}} = D^* + \delta_D + O(\|\Delta\delta\|^2),
	\]
	where \(\delta_N, \delta_D\) are the first-order linear perturbation terms with respect to the noise.
	The error can be written as
	\[
	E(\omega) = \frac{N_{\mathrm{obs}} D^* - N^* D_{\mathrm{obs}}}{D_{\mathrm{obs}} D^*}
	= \frac{\delta_N D^* - N^* \delta_D + O(\|\Delta\delta\|^2)}{D_{\mathrm{obs}} D^*}.
	\]
	We assume that the true signals are well-separated in the latent space, i.e., there exists a positive constant \(\eta > 0\) such that
	\[
	\|\Delta\varepsilon_\varphi^* - \Delta\varepsilon_c^*\| \ge \eta.
	\]

	The well-conditioning assumption directly provides a lower bound for \(D^*\):
	\[
	D^* = \|\Delta\varepsilon_\varphi^* - \Delta\varepsilon_c^*\|^2 \ge \eta^2.
	\]
	For \(D_{\mathrm{obs}}\), the reverse triangle inequality gives:
	\[
	D_{\mathrm{obs}} = \|(\Delta\varepsilon_\varphi^* - \Delta\varepsilon_c^*) + (\Delta\delta_\varphi - \Delta\delta_c)\|^2
	\ge (\|\Delta\varepsilon_\varphi^* - \Delta\varepsilon_c^*\| - \|\Delta\delta_\varphi - \Delta\delta_c\|)^2
	\ge (\eta - \|\Delta\delta_\varphi - \Delta\delta_c\|)^2.
	\]
	Assuming the noise is sufficiently small (e.g., \(\|\Delta\delta_\varphi - \Delta\delta_c\| \le \eta / 2\)), we have
	\[
	D_{\mathrm{obs}} \ge \left(\frac{\eta}{2}\right)^2 = \frac{\eta^2}{4}.
	\]
	Thus, the lower bound for the entire denominator is:
	\[
	D_{\mathrm{obs}} D^* \ge \left(\frac{\eta^2}{4}\right) \cdot \eta^2 = \frac{\eta^4}{4}.
	\]
	
	Since \(\delta_N, \delta_D = O(\|\Delta\delta\|)\), the first-order term in the numerator is bounded:
	\[
	|\delta_N D^* - N^* \delta_D| \le C_1 \|\Delta\delta\|,
	\]
	where \(C_1\) is a constant that depends only on the norms of the true signals.
	
	Combining the upper bound of the numerator and the lower bound of the denominator, we obtain the final error bound:
	\[
	|E(\omega)| \le \frac{C_1 \|\Delta\delta\| + O(\|\Delta\delta\|^2)}{\eta^4 / 4} = \frac{4C_1}{\eta^4} \|\Delta\delta\| + O(\|\Delta\delta\|^2) = O(\|\Delta\delta\|).
	\]
	Therefore, the approximate solution is a first-order Lipschitz continuous map from the noisy input to the output error, meaning the computational error is on the same order as the input noise.
\end{proof}

\section{Generalized POLARIS under Score and Flow based Parameterizations}

Consider the probability flow ODE with field $\psi_\theta(x, t)$. For noise, score, or velocity, the inverse step is $x_i = x_{i+1} - \Delta t \, \psi_\theta(x_{i+1}, t_{i+1})$. POLARIS optimizes $\omega_t$ to minimize the error $\|\tau(t)\|_2^2$.

\subsection{Unified POLARIS Update Rule}

Let $\Delta \psi_\emptyset(t)$ and $\Delta \psi_c(t)$ denote the changes in unconditional and conditional predictions along the inversion trajectory. The error is approximated as $\tau(t) = (1-\omega_t)\Delta \psi_\emptyset(t) + \omega_t \Delta \psi_c(t)$. Minimizing $\|\tau(t)\|_2^2$ yields a unified closed-form update:
\begin{equation}
    \omega_t^* = \frac{\|\Delta \psi_\emptyset\|_2^2 - \Delta \psi_\emptyset^\top \Delta \psi_c}{\|\Delta \psi_\emptyset - \Delta \psi_c\|_2^2}.
    \label{eq:unified-polaris}
\end{equation}
We now instantiate this generic rule for two specific parameterizations:

\paragraph{1. Score-based parameterization.}
The network predicts the score $s_\theta(x,t) \propto -\epsilon_\theta(x,t)/\sigma_t$. Setting $\psi = s$, Eq.~\eqref{eq:unified-polaris} minimizes the error in the \textit{score space}. Since $\sigma_t \to 0$ as $t \to 0$, the metric penalizes errors in the final generation steps.

\paragraph{2. Flow-based parameterization.}
The network predicts velocity $v_\theta(x,t) = \alpha_t \epsilon_\theta(x,t) + \beta_t x_t$. By setting $\psi = v$, Eq.~\eqref{eq:unified-polaris} minimizes the error in the \textit{velocity space}. This space offers a straighter, more numerically stable trajectory for optimization.

\subsection{Invariance of Fixed-Scale Baseline}

While POLARIS yields different trajectories depending on the chosen space (due to the metric change in the norm $\|\cdot\|_2^2$), we prove that the standard fixed-scale CFG baseline remains invariant.

\begin{theorem}
    For a fixed scale $\omega$, the generative step is identical across Noise, Score, and Flow parameterizations.
\end{theorem}

\begin{proof}
    Both Score and Flow relate to the noise $\epsilon$ via a time-dependent affine transformation $\mathcal{T}_t(\cdot)$:
\begin{equation}
    \psi_t = \mathbf{a}_t \epsilon_t + \mathbf{b}_t.
\end{equation}
For Score, $\mathbf{a}_t \propto -1/\sigma_t, \mathbf{b}_t=0$. For Flow, $\mathbf{a}_t = \sqrt{\bar{\alpha}_t}, \mathbf{b}_t = -\sqrt{1-\bar{\alpha}_t}x_t$.
Applying CFG in the transformed space $\psi$:
\begin{align}
    \psi_{\text{guided}} &= \psi_\emptyset + \omega (\psi_c - \psi_\emptyset) \nonumber \\
    (\mathbf{a}_t \epsilon_{\text{guided}} + \mathbf{b}_t) &= (\mathbf{a}_t \epsilon_\emptyset + \mathbf{b}_t) + \omega \left[ (\mathbf{a}_t \epsilon_c + \mathbf{b}_t) - (\mathbf{a}_t \epsilon_\emptyset + \mathbf{b}_t) \right].
\end{align}
The translation term $\mathbf{b}_t$ cancels out in the subtraction. Dividing by the scalar $\mathbf{a}_t$, we recover the noise-space update:
\begin{equation}
    \epsilon_{\text{guided}} = \epsilon_\emptyset + \omega (\epsilon_c - \epsilon_\emptyset).
\end{equation}
Thus, the baseline trajectory is mathematically independent of the parameterization space. 
\end{proof}

\subsection{Experimental Validation}
We evaluate POLARIS on COCO 2017 and Pick-a-Pic (100 samples each) by transforming SD v1.5 outputs into score and velocity spaces. Using 50 DDIM steps and replayed $\omega_t$, Table~\ref{tab:score_flow_comparison} confirms effectiveness across parameterizations, with the Score-based variant achieving the highest fidelity.

\begin{table}[h]
  \centering
  \caption{Quantitative comparison of POLARIS variants under Score and Flow parameterizations.}
  \label{tab:score_flow_comparison}
  \begin{tabular}{l l c c c c}
    \toprule
    \textbf{Dataset} & \textbf{Paradigm} & \textbf{MSE} ($\downarrow$) & \textbf{PSNR} ($\uparrow$) & \textbf{SSIM} ($\uparrow$) & \textbf{LPIPS} ($\downarrow$) \\
    \midrule
    
    \rowcolor{color1} 
    \cellcolor{white}\multirow{2}{*}{COCO 2017} 
      & Score-based & 333.64 & 23.74 & 0.7353 & 0.1402 \\
      & Flow-based  & 389.72 & 23.24 & 0.7234 & 0.1583 \\
    \midrule
    
    \rowcolor{color1} 
    \cellcolor{white}\multirow{2}{*}{Pick-a-Pic} 
      & Score-based & 235.03 & 25.59 & 0.8262 & 0.0758 \\
      & Flow-based  & 311.98 & 25.14 & 0.8193 & 0.0858 \\
    \bottomrule
  \end{tabular}
\end{table}

\section{Additional Experimental Results}
\label{app: additional experimental resuls}

\subsection{Computational Efficiency}
\label{sec:efficiency}

\noindent 
\begin{minipage}[c]{0.58\linewidth} 
    We conduct a rigorous efficiency benchmark in Tab.~\ref{tab:efficiency_comparison} using 50 inference steps on a single NVIDIA RTX 4090. 
    Optimization-based methods like Null-Text Inversion achieve high fidelity but suffer from prohibitive latency, rendering them unsuitable for interactive applications. Similarly, EDICT doubles the computational load. 
    In contrast, POLARIS operates as a lightweight module. It boosts the DDIM baseline PSNR by \textbf{+9.39 dB} while incurring only $\sim$3\% overhead. Notably, it outperforms other optimization-free approaches like Negative-Prompt  and Direct Inversion in speed, offering the best fidelity-latency trade-off.
\end{minipage}
\hfill 
\begin{minipage}[c]{0.40\linewidth} 
    \centering 
    
    \vspace{-3pt} 
    \captionof{table}{\textbf{Efficiency Comparison.}} 
    \label{tab:efficiency_comparison}
    
    \vspace{-8pt} 
    
    \footnotesize 
    \setlength{\tabcolsep}{0pt} 
    \begin{tabular*}{\linewidth}{l@{\extracolsep{\fill}}cc}
        \toprule
        \textbf{Method} & \textbf{PSNR} & \textbf{Time (s)} \\
        \midrule
        DDIM inversion & 14.10 & \textbf{7.71} \\
        Null-Text inversion & 26.11 & 135.48 \\
        Negative-prompt inversion & 23.38 & 8.32 \\
        EDICT & \textbf{26.53} & 16.21 \\
        Direct inversion & \underline{26.32} & 12.28 \\
        \midrule
        \textbf{POLARIS(Ours)} & 23.49 & \underline{7.96} \\
        \bottomrule
    \end{tabular*}
\end{minipage}

\vspace{10pt} 

\noindent\textbf{Scalability analysis.} To further validate the negligible overhead of POLARIS, we report the runtime across varying inference steps (from 10 to 100) in Tab.~\ref{tab:computation_cost_detail}. As shown, the additional time introduced by POLARIS averages only \textbf{3.3\%} compared to the fixed-scale baseline, confirming its scalability and efficiency.

\begin{table}[h!]
    \centering
    \caption{Runtime comparison (s) between Fixed Scale and POLARIS across different inference steps.}
    \label{tab:computation_cost_detail}
    
    \small 
    
    \begin{tabular*}{\linewidth}{l@{\extracolsep{\fill}}cccccccccc}
        \toprule
        \textbf{Method} & \textbf{10} & \textbf{20} & \textbf{30} & \textbf{40} & \textbf{50} & \textbf{60} & \textbf{70} & \textbf{80} & \textbf{90} & \textbf{100} \\
        \midrule
Fixed scale & 1.6876 & 3.1628 & 4.6906 & 6.2423 & 7.7075 & 9.2745 & 10.5553 & 12.3849 & 13.7852 & 15.0787 \\
POLARIS     & 1.7113 & 3.2435 & 4.8286 & 6.3940 & 7.9618 & 9.5557 & 11.1370 & 12.6824 & 14.3093 & 15.9283 \\
        \bottomrule
    \end{tabular*}
\end{table}

\subsection{Image Reconstruction}
As a control experiment, we further conducted reconstruction tests under standard DDIM, 
where both inversion and sampling guidance scales were fixed to $\omega \equiv 1$. 
Using the same Stable Diffusion v1.5 backbone, we evaluated reconstruction performance 
from 10 to 100 inference steps at intervals of 10. 
The results show that this baseline consistently underperforms POLARIS across all metrics and step counts.

\begin{table}[h!]
\centering
\caption{Reconstruction Metrics Comparison on COCO and Pick-a-Pic Datasets with reconstruction scale $=1$ as contrast.}
\label{tab:combined_metrics_moderate}
\begin{tabular}{@{}llcccccccccc@{}}
\toprule
\textbf{Metric} & \textbf{Dataset} & \textbf{10} & \textbf{20} & \textbf{30} & \textbf{40} & \textbf{50} & \textbf{60} & \textbf{70} & \textbf{80} & \textbf{90} & \textbf{100} \\
\midrule
\multirow{2}{*}{MSE ($\downarrow$)}
  & COCO2017         & 621.45 & 612.81 & 605.93 & 594.17 & 585.22 & 573.99 & 568.14 & 560.29 & 555.71 & 552.01 \\
  & Pick-a-Pic   & 672.81 & 661.90 & 650.15 & 638.49 & 627.31 & 615.82 & 610.93 & 604.18 & 598.22 & 594.19 \\
\midrule
\multirow{2}{*}{PSNR ($\uparrow$)}
  & COCO2017         & 19.85  & 19.98  & 20.07  & 20.21  & 20.35  & 20.48  & 20.59  & 20.68  & 20.75  & 20.81  \\
  & Pick-a-Pic   & 18.99  & 19.11  & 19.23  & 19.35  & 19.47  & 19.59  & 19.66  & 19.74  & 19.82  & 19.88  \\
\midrule
\multirow{2}{*}{SSIM ($\uparrow$)}
  & COCO2017         & 0.5011 & 0.5188 & 0.5290 & 0.5381 & 0.5453 & 0.5512 & 0.5549 & 0.5580 & 0.5601 & 0.5615 \\
  & Pick-a-Pic   & 0.4123 & 0.4297 & 0.4411 & 0.4503 & 0.4588 & 0.4654 & 0.4688 & 0.4719 & 0.4740 & 0.4752 \\
\midrule
\multirow{2}{*}{LPIPS ($\downarrow$)}
  & COCO2017         & 0.2144 & 0.1981 & 0.1905 & 0.1842 & 0.1798 & 0.1743 & 0.1712 & 0.1689 & 0.1670 & 0.1659 \\
  & Pick-a-Pic   & 0.2691 & 0.2553 & 0.2478 & 0.2411 & 0.2350 & 0.2298 & 0.2265 & 0.2231 & 0.2209 & 0.2190 \\
\bottomrule
\end{tabular}
\end{table}

\subsection{Ablations on  Sampling Schedulers}
Table~\ref{tab:metric_combined_strategies} compares POLARIS with four alternative guidance schedules on COCO2017 using Stable Diffusion~v1.5 under identical settings: the forward schedule ($\omega_t$), reverse schedule ($\omega_{T-t+1}$), cosine decay (\textsc{Cosine}\,$\downarrow$) from $1$ to $0$, and a fixed-scale baseline ($\omega\!\equiv\!1$). Across 10 - 100 inference steps, POLARIS consistently attains the lowest MSE and LPIPS and the highest PSNR and SSIM, indicating superior reconstruction fidelity and structural preservation, while the fixed-scale baseline performs substantially worse, underscoring the importance of dynamic guidance scaling in both inversion and sampling.

\begin{table*}[h!]
  \centering
  \caption{Performance comparison of POLARIS's and other sampling strategies across various inference steps on COCO2017.}
  \label{tab:metric_combined_strategies}
  \begin{adjustbox}{max width=\textwidth}
  \begin{tabular}{l ll *{10}{c}}
    \toprule
    \textbf{Dataset} & \textbf{Metric} & \textbf{Schedulers} & \textbf{10} & \textbf{20} & \textbf{30} & \textbf{40} & \textbf{50} & \textbf{60} & \textbf{70} & \textbf{80} & \textbf{90} & \textbf{100} \\
    \midrule

\multirow{16}{*}{\textbf{COCO2017}} 
    & \multirow{4}{*}{MSE ($\downarrow$)} & \cellcolor{color1}$\omega_{T-t+1}$ &
      \cellcolor{color1}506 & \cellcolor{color1}498 & \cellcolor{color1}490 & \cellcolor{color1}493 & \cellcolor{color1}498 &
      \cellcolor{color1}467 & \cellcolor{color1}477 & \cellcolor{color1}464 & \cellcolor{color1}478 & \cellcolor{color1}476 \\
    & & $\omega_t$ &
      688 & 651 & 635 & 639 & 642 & 610 & 615 & 608 & 619 & 617 \\
    & &$\omega\equiv1$ &
      1254 & 1198 & 1153 & 1160 & 1171 & 1120 & 1135 & 1115 & 1142 & 1138 \\
    & & \textsc{Cosine↓} &
      891 & 845 & 822 & 829 & 833 & 798 & 804 & 790 & 810 & 805 \\
    \addlinespace[0.6ex]
    & \multirow{4}{*}{PSNR  ($\uparrow$)} & \cellcolor{color1}$\omega_{T-t+1}$ &
      \cellcolor{color1}22.04 & \cellcolor{color1}22.24 & \cellcolor{color1}22.36 & \cellcolor{color1}22.36 & \cellcolor{color1}22.34 &
      \cellcolor{color1}22.62 & \cellcolor{color1}22.54 & \cellcolor{color1}22.66 & \cellcolor{color1}22.53 & \cellcolor{color1}22.55 \\
    & & $\omega_t$ &
      20.75 & 20.99 & 21.11 & 21.08 & 21.05 & 21.28 & 21.23 & 21.30 & 21.21 & 21.22 \\
    & & $\omega\equiv1$ &
      18.15 & 18.34 & 18.51 & 18.48 & 18.43 & 18.64 & 18.58 & 18.66 & 18.55 & 18.57 \\
    & & \textsc{Cosine↓} &
      19.63 & 19.86 & 19.98 & 19.95 & 19.92 & 20.11 & 20.07 & 20.16 & 20.04 & 20.06 \\
    \addlinespace[0.6ex]
    & \multirow{4}{*}{SSIM  ($\uparrow$)} & \cellcolor{color1}$\omega_{T-t+1}$ &
      \cellcolor{color1}0.6922 & \cellcolor{color1}0.7043 & \cellcolor{color1}0.7085 & \cellcolor{color1}0.7087 & \cellcolor{color1}0.7088 &
      \cellcolor{color1}0.7135 & \cellcolor{color1}0.7122 & \cellcolor{color1}0.7143 & \cellcolor{color1}0.7126 & \cellcolor{color1}0.7133 \\
    & & $\omega_t$ &
      0.6615 & 0.6730 & 0.6781 & 0.6779 & 0.6780 & 0.6831 & 0.6820 & 0.6845 & 0.6822 & 0.6830 \\
    & & $\omega\equiv1$ &
      0.5890 & 0.5995 & 0.6044 & 0.6039 & 0.6041 & 0.6088 & 0.6075 & 0.6101 & 0.6077 & 0.6085 \\
    & & \textsc{Cosine↓} &
      0.6251 & 0.6368 & 0.6415 & 0.6410 & 0.6412 & 0.6466 & 0.6452 & 0.6480 & 0.6455 & 0.6463 \\
    \addlinespace[0.6ex]
    & \multirow{4}{*}{LPIPS ($\downarrow$)} & \cellcolor{color1}$\omega_{T-t+1}$ &
      \cellcolor{color1}0.2231 & \cellcolor{color1}0.2039 & \cellcolor{color1}0.1957 & \cellcolor{color1}0.1950 & \cellcolor{color1}0.1955 &
      \cellcolor{color1}0.1870 & \cellcolor{color1}0.1897 & \cellcolor{color1}0.1862 & \cellcolor{color1}0.1892 & \cellcolor{color1}0.1886 \\
    & & $\omega_t$ &
      0.2685 & 0.2451 & 0.2358 & 0.2349 & 0.2352 & 0.2255 & 0.2281 & 0.2240 & 0.2275 & 0.2268 \\
    & & $\omega\equiv1$ &
      0.4102 & 0.3855 & 0.3710 & 0.3701 & 0.3715 & 0.3601 & 0.3640 & 0.3588 & 0.3633 & 0.3621 \\
    & & \textsc{Cosine↓} &
      0.3350 & 0.3106 & 0.2988 & 0.2975 & 0.2981 & 0.2870 & 0.2905 & 0.2855 & 0.2899 & 0.2888 \\
    \addlinespace[0.8ex]
    \midrule
  \end{tabular}
  \end{adjustbox}
\end{table*}

\subsection{Scaling POLARIS to Larger Diffusion Models}
We further extend POLARIS to \emph{Stable Diffusion XL}, demonstrating that our method is not restricted to SD~1.5 but also scales effectively to larger, higher-capacity diffusion models. As shown in Table~\ref{tab:sdxl}, POLARIS achieves even better results on this larger-parameter backbone, further confirming its scalability to large scale models.

\begin{table*}[h]
  \centering
  \caption{Performance comparison of POLARIS and Fixed scale across various inference steps on Pick-a-Pic and COCO2017 in SDXL.}
  \label{tab:sdxl}
  \begin{adjustbox}{max width=\textwidth}
  \begin{tabular}{l ll *{10}{c}}
    \toprule
    \textbf{Dataset} & \textbf{Metric} & \textbf{Method} & \textbf{10} & \textbf{20} & \textbf{30} & \textbf{40} & \textbf{50} & \textbf{60} & \textbf{70} & \textbf{80} & \textbf{90} & \textbf{100} \\
    \midrule

\multirow{8}{*}{\textbf{COCO2017}} 
    & \multirow{2}{*}{MSE ($\downarrow$)} & Fixed scale &
      1558 & 1932 & 2092 & 2348 & 2436 & 2013 & 2316 & 2071 & 2516 & 2668 \\
    & & \cellcolor{color1}POLARIS &
      \cellcolor{color1}479 & \cellcolor{color1}435 & \cellcolor{color1}401 & \cellcolor{color1}385 & \cellcolor{color1}366 &
      \cellcolor{color1}292 & \cellcolor{color1}310 & \cellcolor{color1}258 & \cellcolor{color1}274 & \cellcolor{color1}283 \\
    \addlinespace[0.6ex]
    & \multirow{2}{*}{PSNR  ($\uparrow$)} & Fixed scale &
      16.52 & 15.58 & 15.27 & 14.76 & 14.60 & 15.43 & 14.82 & 15.32 & 14.45 & 14.19 \\
    & & \cellcolor{color1}POLARIS &
      \cellcolor{color1}21.97 & \cellcolor{color1}22.54 & \cellcolor{color1}23.03 & \cellcolor{color1}23.34 & \cellcolor{color1}23.61 &
      \cellcolor{color1}24.64 & \cellcolor{color1}24.49 & \cellcolor{color1}25.18 & \cellcolor{color1}25.03 & \cellcolor{color1}24.85 \\
    \addlinespace[0.6ex]
    & \multirow{2}{*}{SSIM  ($\uparrow$)} & Fixed scale &
      0.5950 & 0.5856 & 0.5744 & 0.5575 & 0.5500 & 0.5711 & 0.5527 & 0.5661 & 0.5405 & 0.5322 \\
    & & \cellcolor{color1}POLARIS &
      \cellcolor{color1}0.7113 & \cellcolor{color1}0.7478 & \cellcolor{color1}0.7574 & \cellcolor{color1}0.7651 & \cellcolor{color1}0.7698 &
      \cellcolor{color1}0.7854 & \cellcolor{color1}0.7828 & \cellcolor{color1}0.7967 & \cellcolor{color1}0.7950 & \cellcolor{color1}0.7933 \\
    \addlinespace[0.6ex]
    & \multirow{2}{*}{LPIPS ($\downarrow$)} & Fixed scale &
      0.4636 & 0.4914 & 0.5050 & 0.5205 & 0.5279 & 0.5004 & 0.5207 & 0.5053 & 0.5354 & 0.5467 \\
    & & \cellcolor{color1}POLARIS &
      \cellcolor{color1}0.2832 & \cellcolor{color1}0.2568 & \cellcolor{color1}0.2489 & \cellcolor{color1}0.2366 & \cellcolor{color1}0.2293 &
      \cellcolor{color1}0.2021 & \cellcolor{color1}0.2054 & \cellcolor{color1}0.1815 & \cellcolor{color1}0.1840 & \cellcolor{color1}0.1872 \\
        \addlinespace[0.8ex]
    \midrule
    \multirow{8}{*}{\textbf{Pick-a-Pic}} 
    & \multirow{2}{*}{MSE ($\downarrow$)} & Fixed scale &
      1149 & 1499 & 1604 & 1842 & 1928 & 1643 & 1868 & 1742 & 2068 & 2214 \\
    & & \cellcolor{color1}POLARIS &
      \cellcolor{color1}309 & \cellcolor{color1}342 & \cellcolor{color1}326 & \cellcolor{color1}337 & \cellcolor{color1}326 &
      \cellcolor{color1}269 & \cellcolor{color1}268 & \cellcolor{color1}255 & \cellcolor{color1}252 & \cellcolor{color1}240 \\
    \addlinespace[0.6ex]
    & \multirow{2}{*}{PSNR ($\uparrow$)} & Fixed scale &
      18.07 & 16.84 & 16.60 & 16.01 & 15.81 & 16.53 & 15.96 & 16.26 & 15.49 & 15.21 \\
    & & \cellcolor{color1}POLARIS &
      \cellcolor{color1}24.01 & \cellcolor{color1}24.22 & \cellcolor{color1}24.59 & \cellcolor{color1}24.60 & \cellcolor{color1}24.83 &
      \cellcolor{color1}25.88 & \cellcolor{color1}25.88 & \cellcolor{color1}26.38 & \cellcolor{color1}25.98 & \cellcolor{color1}26.30 \\
    \addlinespace[0.6ex]
    & \multirow{2}{*}{SSIM ($\uparrow$)} & Fixed scale &
      0.6884 & 0.6735 & 0.6695 & 0.6530 & 0.6455 & 0.6628 & 0.6462 & 0.6538 & 0.6305 & 0.6222 \\
    & & \cellcolor{color1}POLARIS &
      \cellcolor{color1}0.8014 & \cellcolor{color1}0.8181 & \cellcolor{color1}0.8297 & \cellcolor{color1}0.8323 & \cellcolor{color1}0.8359 &
      \cellcolor{color1}0.8478 & \cellcolor{color1}0.8523 & \cellcolor{color1}0.8552 & \cellcolor{color1}0.8579 & \cellcolor{color1}0.8625 \\
    \addlinespace[0.6ex]
    & \multirow{2}{*}{LPIPS ($\downarrow$)} & Fixed scale &
      0.3306 & 0.3745 & 0.3778 & 0.4002 & 0.4080 & 0.3846 & 0.4032 & 0.3928 & 0.4247 & 0.4360 \\
    & & \cellcolor{color1}POLARIS &
      \cellcolor{color1}0.1673 & \cellcolor{color1}0.1745 & \cellcolor{color1}0.1641 & \cellcolor{color1}0.1636 & \cellcolor{color1}0.1557 &
      \cellcolor{color1}0.1353 & \cellcolor{color1}0.1298 & \cellcolor{color1}0.1241 & \cellcolor{color1}0.1226 & \cellcolor{color1}0.1150 \\

    \addlinespace[0.8ex]
    
    \bottomrule
  \end{tabular}
  \end{adjustbox}
\end{table*}

\section{Experimental Details}
\subsection{Baselines}
\label{baselines}
\noindent\textbf{DDRM. \cite{kawar2022denoising}} Denoising Diffusion Restoration Models formulates a solution to general linear inverse problems without requiring task-specific training. The key insight is to treat the pre-trained diffusion model as a powerful generative prior. At each reverse diffusion step, it first predicts a clean image and then applies a correction term derived from the measurement error. This iterative projection ensures that the final output is consistent with the given measurements, such as a blurred or downsampled image. We use a correction strength of $\eta=0.1$.\\

\noindent\textbf{DDNM. \cite{wang2022zero}} Denoising Diffusion Null-space Models offers a highly efficient, analytical approach by leveraging the properties of linear transformations. It decomposes the problem space into two orthogonal subspaces: the range-space, where the measurement provides deterministic information, and the null-space, where information is lost. DDNM uses the measurement to perfectly reconstruct the range-space component and employs the diffusion model exclusively to generate plausible details for the null-space component. This separation prevents the model from hallucinating artifacts in data-constrained regions.\\

\noindent\textbf{DPS. \cite{chung2024diffusionposteriorsamplinggeneral}} Diffusion Posterior Sampling introduces a flexible, gradient-based guidance mechanism applicable to both linear and non-linear inverse problems. It interprets the reverse diffusion process as a form of posterior sampling and guides the sampling trajectory using the gradient of a likelihood function. In practice, this is implemented by computing the gradient of a consistency loss function that measures how well the current prediction satisfies the problem's constraints. This gradient is then used to perturb the denoised sample at each step, steering it towards a high-likelihood solution. The guidance strength is set to $\lambda=0.2$.\\

\noindent\textbf{P2P. \cite{hertz2022prompt}} Prompt-to-Prompt is a seminal zero-shot method for text-guided image editing that requires no model training or fine-tuning. Its key insight is that the cross-attention maps generated during the denoising process implicitly control the spatial layout and structure of the synthesized image. The method enables editing by injecting the attention maps from a source image's generation process into the generation process of a target prompt. This ensures that the resulting image retains the structure of the original while incorporating the semantic changes from the new prompt.\\

\noindent\textbf{P2P Mask.} We introduce P2P-Mask, a dual-pipeline approach for high-fidelity, zero-shot real image editing. The process initiates with image inversion to encode the source image into its initial noisy latent representation, for which we compare two strategies: a standard fixed scale DDIM inversion and our POLARIS method, which calculates an optimal guidance weight at each step. In the subsequent spatially-aware generation phase, we reconstruct the image using the target prompt. The core of this phase leverages cross-attention maps to dynamically generate a soft spatial mask at each denoising step by analyzing the relative attention between 'edit tokens' and 'preserve tokens'. This mask is then used to spatially modulate the guidance strength, applying strong creative guidance to the edit regions while imposing a matching reconstruction scale, correspondingly fixed or dynamic based on the inversion method, to the background. This design ensures that edits are precisely confined to the target area while maximizing the reconstruction fidelity of non-edited regions.\\

\noindent\textbf{SAGE. \cite{Gomez_Trenado_2025}} Self-Attention Guidance for image Editing is a novel method for text-guided image editing that introduces a guidance mechanism based on self-attention maps. The core insight is that the self-attention maps generated during the inverse process contain rich structural and contextual information about the source image. SAGE leverages this by first performing DDIM inversion with the source prompt ($c_{s}$) to record these intermediate self-attention maps. Subsequently, during the forward sampling process with the target prompt ($c^{t}$), it enforces consistency by minimizing a loss between the newly generated self-attention maps and the recorded ones. This allows for high-fidelity preservation of unedited regions without requiring explicit or costly reconstruction steps, striking an effective balance between editing accuracy and structural integrity.\\

\noindent\textbf{Stable Diffusion v1.5. \cite{rombach2022highresolutionimagesynthesislatent}} All of our experiments are built upon Stable Diffusion v1.5, a powerful, publicly available latent diffusion model. It operates in a compressed latent space for computational efficiency and uses a frozen CLIP ViT-L/14 text encoder to interpret text prompts. The model's UNet architecture is conditioned on these text embeddings via cross-attention, enabling it to generate high-fidelity images from a vast range of textual descriptions. We use the official 'v1-5' checkpoint as the foundational backbone for all compared methods.\\

\newpage

\noindent\textbf{Stable Diffusion XL.} SDXL represents a significant evolution in latent diffusion models, featuring a substantially larger UNet architecture with approximately 2.6 billion parameters three times the size of SD v1.5. Unlike its predecessor, SDXL employs a dual-text encoder system, combining CLIP ViT-L and OpenCLIP ViT-G/14, to capture both high level semantics and fine-grained details. It is trained natively at a $1024 \times 1024$ resolution and utilizes micro-conditioning to handle varying aspect ratios. We incorporate SDXL into our experiments to validate the scalability of POLARIS and its effectiveness on sota, high-capacity foundation models.\\ 

\noindent\textbf{Null-Text Inversion. \cite{mokady2022null}} It is an optimization-based technique designed to enable accurate reconstruction for real image editing. Recognizing that CFG amplifies inversion errors, it modifies the unconditional textual embedding (the ''null text'') rather than the latent code. At each timestep of the inversion process, it performs an iterative optimization to find a specific null embedding that steers the denoising trajectory back to the original latent state. This allows for high-fidelity reconstruction but incurs a significant computational cost due to the per-step optimization loop.\\

\noindent\textbf{Negative-Prompt Inversion. \cite{miyake2024negativepromptinversionfastimage}} It offers a more efficient alternative to heavy optimization methods. Instead of optimizing the null embedding at every single timestep independently, it seeks to optimize the negative prompt embedding to absorb the inversion error. By shifting the discrepancy caused by CFG into the negative prompt space, it allows the forward process to reconstruct the image using the standard sampling mechanism. This approach aims to maintain the editability of the positive prompt while correcting the trajectory drift, often achieving a better balance between reconstruction quality and inference speed compared to full per-step optimization.\\

\noindent\textbf{Direct Inversion. \cite{ju2023directinversionboostingdiffusionbased}} It proposes a straightforward, one-shot solution to the reconstruction problem without iterative optimization. It acknowledges that the standard DDIM inversion path and the generation path diverge due to CFG. To bridge this gap, it explicitly records the difference (the error maps) between the inversion noise predictions and the reconstruction requirements at each step. During the editing phase, these recorded error maps are injected back into the diffusion process, forcing the trajectory to align with the source image's spatial layout while allowing the target prompt to alter the semantics.\\

\noindent\textbf{EDICT. \cite{meng2022sdedit}} Exact Diffusion Inversion via Coupled Transformations fundamentally resolves the non-invertibility of standard diffusion steps by redesigning the sampling process. It employs a dual-variable formulation, maintaining two coupled latent states that are updated in an alternating fashion using affine coupling layers. This architecture renders the diffusion process mathematically bijective, enabling theoretically perfect reconstruction of the source image without any optimization or approximation, albeit at the cost of doubling the memory and computational requirements for maintaining the dual states.

\subsection{Datasets}

Our method is evaluated on two standard and challenging benchmarks to assess its performance across diverse scenarios.\\

\noindent\textbf{MS-COCO 2017. \cite{lin2015microsoftcococommonobjects}} The Microsoft Common Objects in Context 2017 validation set is a large-scale dataset widely used for benchmarking image generation and understanding tasks. It features diverse and complex everyday scenes, with each image annotated with five distinct, human-generated captions. This rich descriptive variance makes it an ideal benchmark for evaluating the robustness and generalization capabilities of text-guided image editing models, testing their ability to parse complex prompts and preserve background details.\\

\noindent\textbf{Pick-a-Pic. \cite{kirstain2023pickapicopendatasetuser}} Pick-a-Pic is a large-scale dataset specifically designed to benchmark text-to-image models based on human preferences. It contains a vast collection of text prompts, each paired with multiple generated image candidates, from which human annotators have selected the one that best aligns with the prompt's intent and aesthetic quality. We utilize this dataset to evaluate our method's ability to produce high-quality and semantically accurate edits that align with user expectations, providing a more nuanced assessment of performance beyond automated metrics.\\

\subsection{Metrics}
To quantitatively evaluate the performance of our method and the baselines, we employ a comprehensive set of metrics that assess different aspects of image quality, including pixel-level fidelity, perceptual similarity, and aesthetic appeal. Given a ground truth image $I$ and a generated image $K$, both of size $m \times n$, the metrics are defined as follows:
\\
\noindent\textbf{MSE.} 
Mean Squared Error is a fundamental metric for pixel-level fidelity. It computes the average squared difference between corresponding pixel values. All calculations are performed in the 8-bit integer domain [0, 255]. A lower MSE value signifies a more accurate pixel-wise reconstruction.
\begin{equation}
    \text{MSE} = \frac{1}{mn} \sum_{i=0}^{m-1} \sum_{j=0}^{n-1} [I(i,j) - K(i,j)]^2
\end{equation}\\

\noindent\textbf{PSNR. \cite{fardo2016formalevaluationpsnrquality}} 
Peak Signal-to-Noise Ratio is a widely used metric for reconstruction quality, defined via the MSE. It measures the ratio between the maximum possible power of a signal and the power of the corrupting noise. Expressed on a logarithmic decibel (dB) scale, a higher PSNR value indicates a higher-quality reconstruction.
\begin{equation}
    \text{PSNR} = 10 \cdot \log_{10} \left( \frac{\text{MAX}_I^2}{\text{MSE}} \right)
\end{equation}
where $\text{MAX}_I$ is the maximum possible pixel value, which is 255 for 8-bit images.\\

\noindent\textbf{SSIM. \cite{1284395}} 
Structural Similarity Index Measure evaluates perceptual similarity by assessing the degradation of structural information. Unlike pixel-based metrics, SSIM compares local patterns based on luminance, contrast, and structure. For two image windows $x$ and $y$, it is defined as:
\begin{equation}
    \text{SSIM}(x, y) = \frac{(2\mu_x\mu_y + C_1)(2\sigma_{xy} + C_2)}{(\mu_x^2 + \mu_y^2 + C_1)(\sigma_x^2 + \sigma_y^2 + C_2)}
\end{equation}
where $\mu$ is the mean, $\sigma^2$ is the variance, $\sigma_{xy}$ is the covariance, and $C_1, C_2$ are stabilization constants. The final SSIM is the mean over all windows. A value closer to 1 indicates perfect structural similarity.\\

\noindent\textbf{LPIPS. \cite{zhang2018unreasonableeffectivenessdeepfeatures}} 
Learned Perceptual Image Patch Similarity is an excellent perceptual metric that better correlates with human judgment. It computes the distance between deep features of two image patches, the test image $x$ and the original image $x_0$, extracted from a pre-trained deep neural network. The distance is calculated as:
\begin{equation}
    \text{LPIPS}(x, x_0) = \sum_{l} \frac{w_l}{H_l W_l} \sum_{h,w} \left\| \hat{y}_{hw}^l - \hat{y}_{0,hw}^l \right\|_2^2
\end{equation}
where $\hat{y}^l, \hat{y}_0^l$ are the unit-normalized feature activations from layer $l$ of the network, scaled by layer-specific weights $w_l$. A lower LPIPS score indicates that two images are more perceptually similar.\\

\noindent\textbf{AES. \cite{schuhmann2022laion5bopenlargescaledataset}} 
The Aesthetic Score is used to evaluate the overall visual appeal of the generated images. This score is derived from a pre-trained model, often trained on large-scale datasets with human aesthetic ratings. Unlike fidelity metrics, AES quantifies the subjective quality and artistic merit of the output. A higher score suggests a more aesthetically pleasing and visually coherent image.\\

\noindent\textbf{Preference rate.} 
To calculate the preference rate, we presented participants with the source instruction and a randomized pair of outputs (ours vs. baseline). Users were asked to select the result with better visual fidelity and instruction alignment.
\begin{tcolorbox}[
    colback=gray!10,       
    colframe=gray!10,      
    boxrule=0pt,           
    arc=2pt,               
    left=3mm, right=3mm, top=2mm, bottom=2mm 
]
    \textbf{\textcolor{red!70!black}{[Question]:}} 
    Please compare the two edited images below based on the instruction. \texttt{<image\_pairs.png>}
    
    \vspace{0.5em} 
    
    \begin{enumerate}[label=(\arabic*), leftmargin=3.5em, itemsep=0.2em]
        \item Image A has better visual fidelity and instruction alignment.
        \item Image B has better visual fidelity and instruction alignment.
        \item Both images are of similar quality / Hard to tell.
    \end{enumerate}
\end{tcolorbox}

\section{Additional Visualizations}
\begin{figure}[h]
    \centering
    \includegraphics[width=\linewidth, height=0.87\textheight, keepaspectratio]{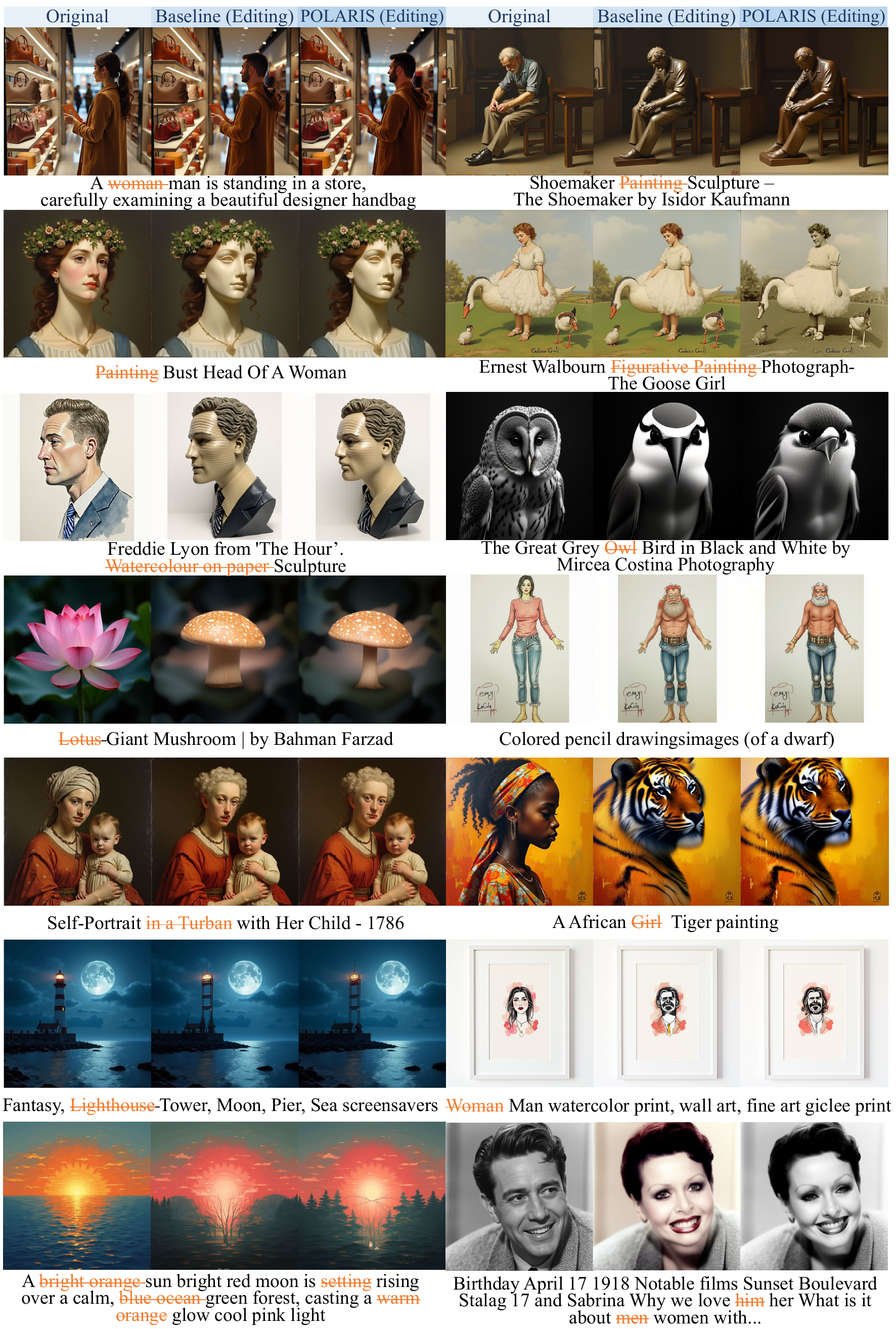}
    
    \caption{\textbf{Editing Visualization.}}
    \label{fig:editing_viz}
\end{figure}

\begin{figure}[t]
    \centering
    \includegraphics[width=\linewidth, height=0.99\textheight, keepaspectratio]{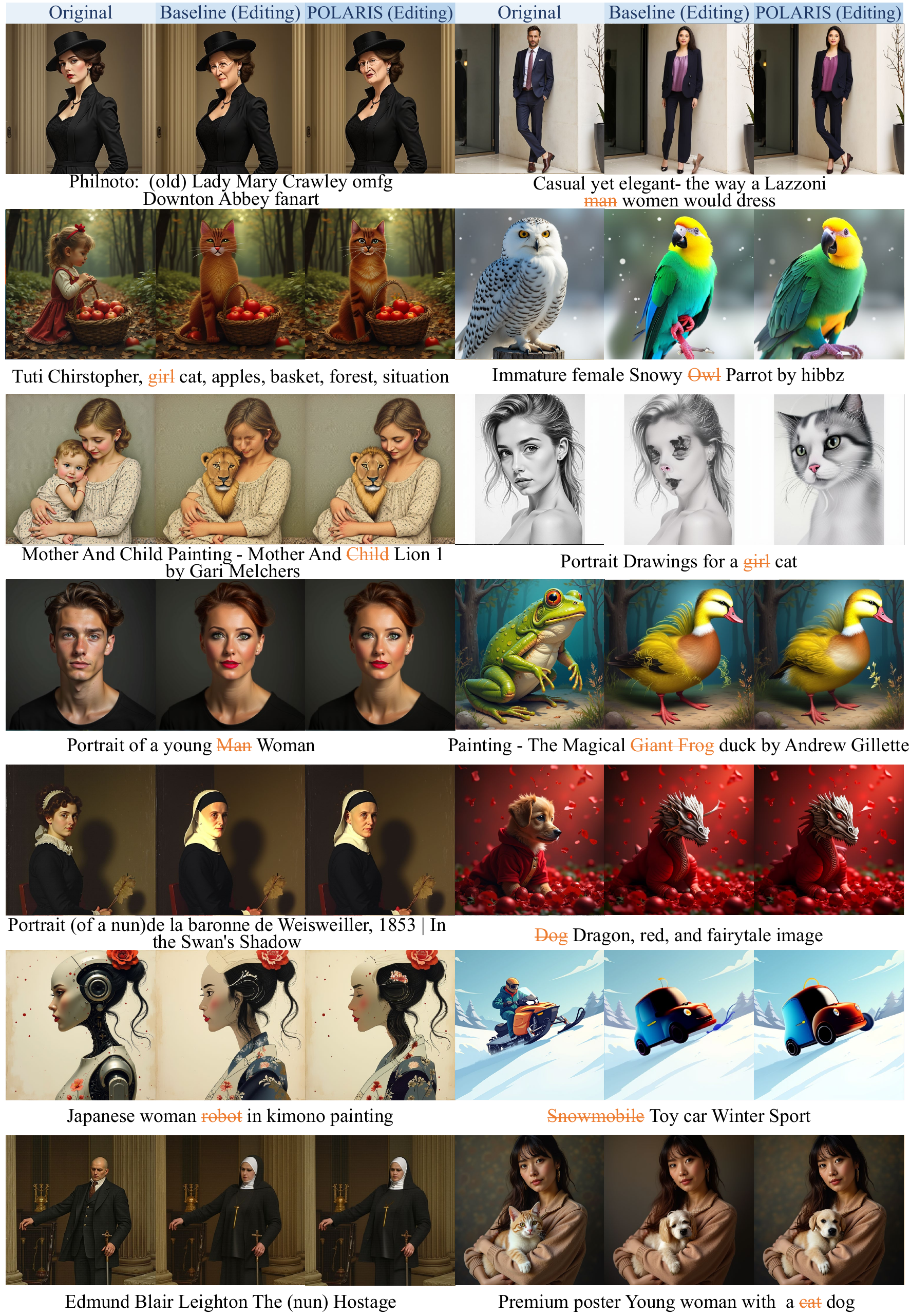}
    
    \caption{\textbf{Editing Visualization.}}
    \label{fig:editing_viz2}
\end{figure}

\begin{figure}[h!]
    \centering
    \includegraphics[width=1.0\linewidth]{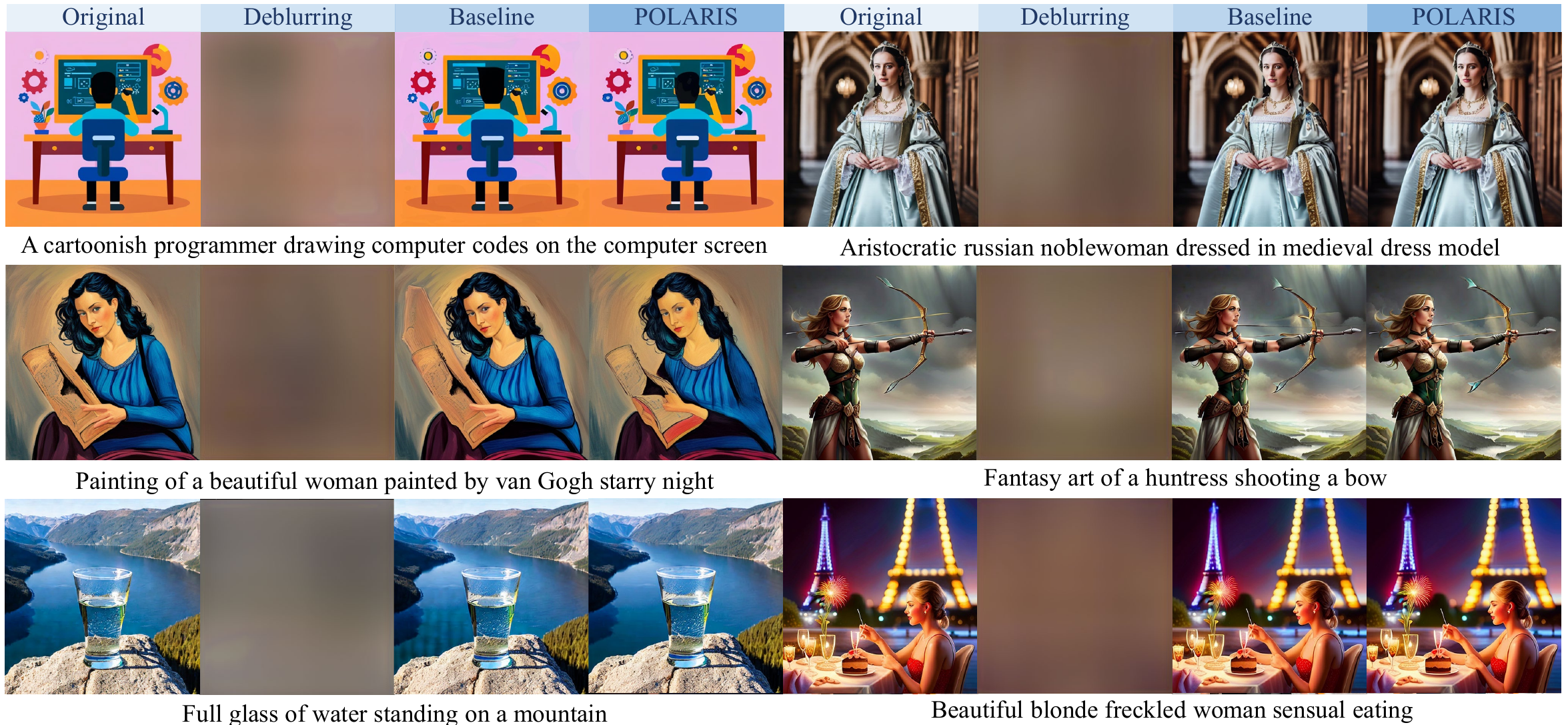}
    \caption{\textbf{Gaussian Deblurring} visualizations }
    \label{fig:supp_deblurring}
\end{figure}

\begin{figure}[h!]
    \centering
    \includegraphics[width=1.0\linewidth]{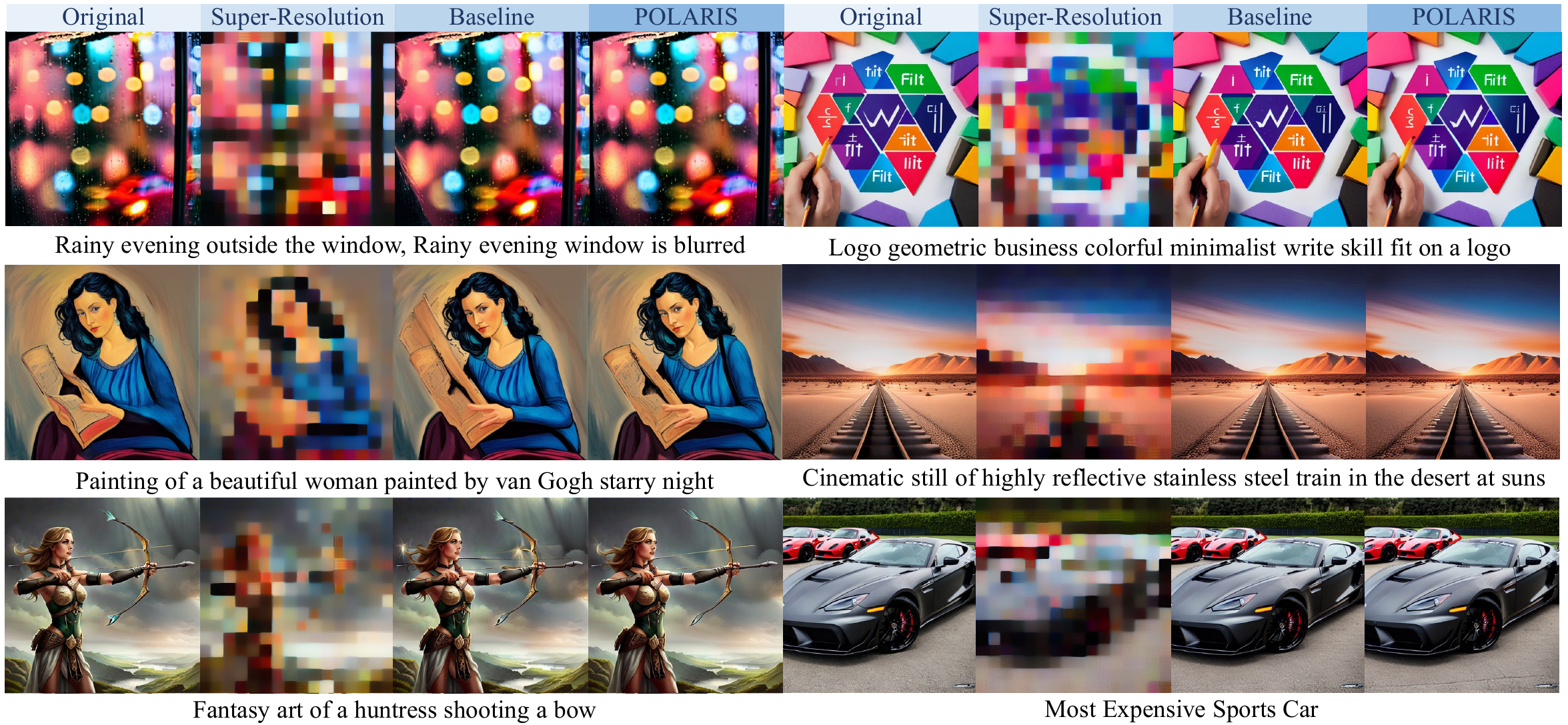}
    \caption{\textbf{$8\times$ Super-Resolution} visualizations}
    \label{fig:supp_sr}
\end{figure}

\begin{figure}[h!]
    \centering
    \includegraphics[width=1.0\linewidth]{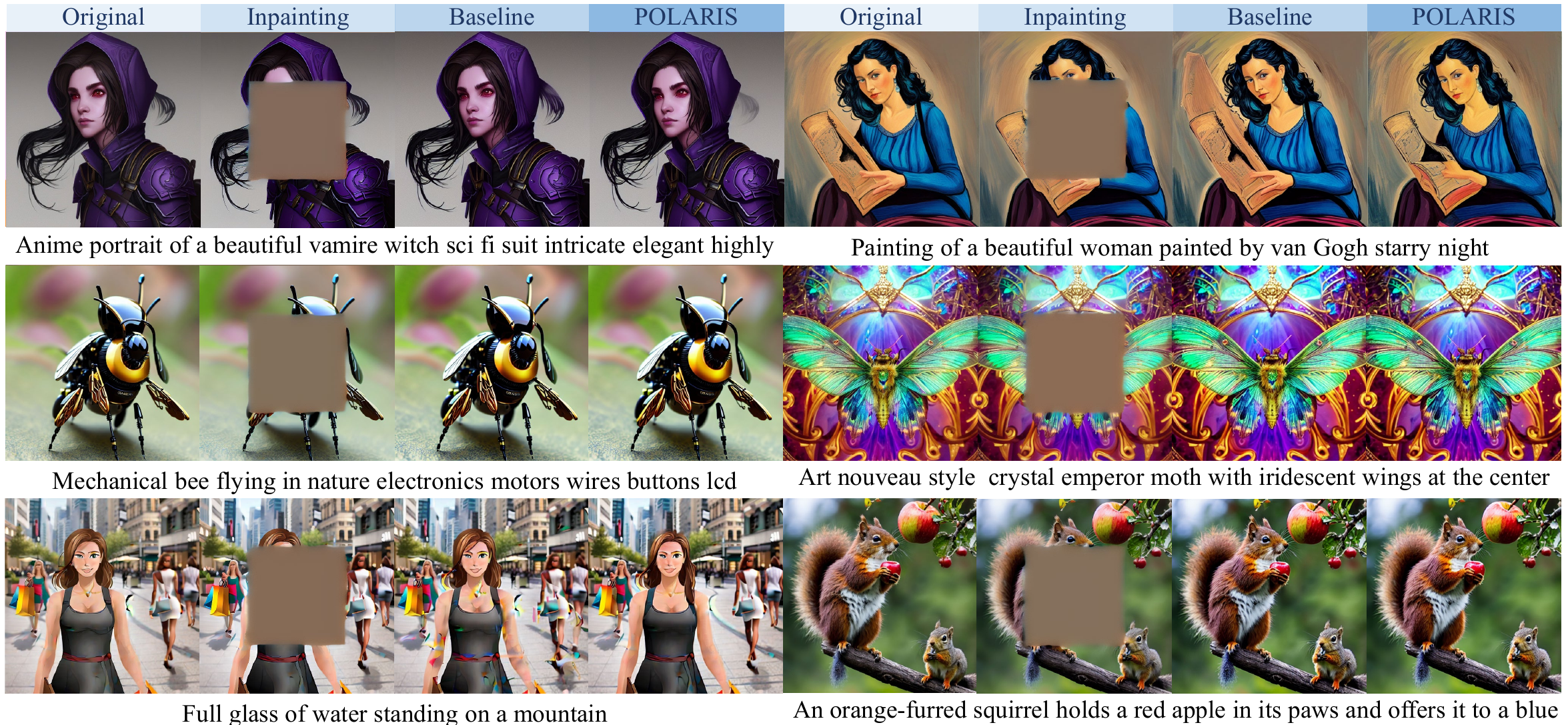}
    \caption{\textbf{Central-mask Inpainting} visualizations}
    \label{fig:supp_inpainting}
\end{figure}

\begin{figure}[h!]
    \centering
    \includegraphics[width=1.0\linewidth]{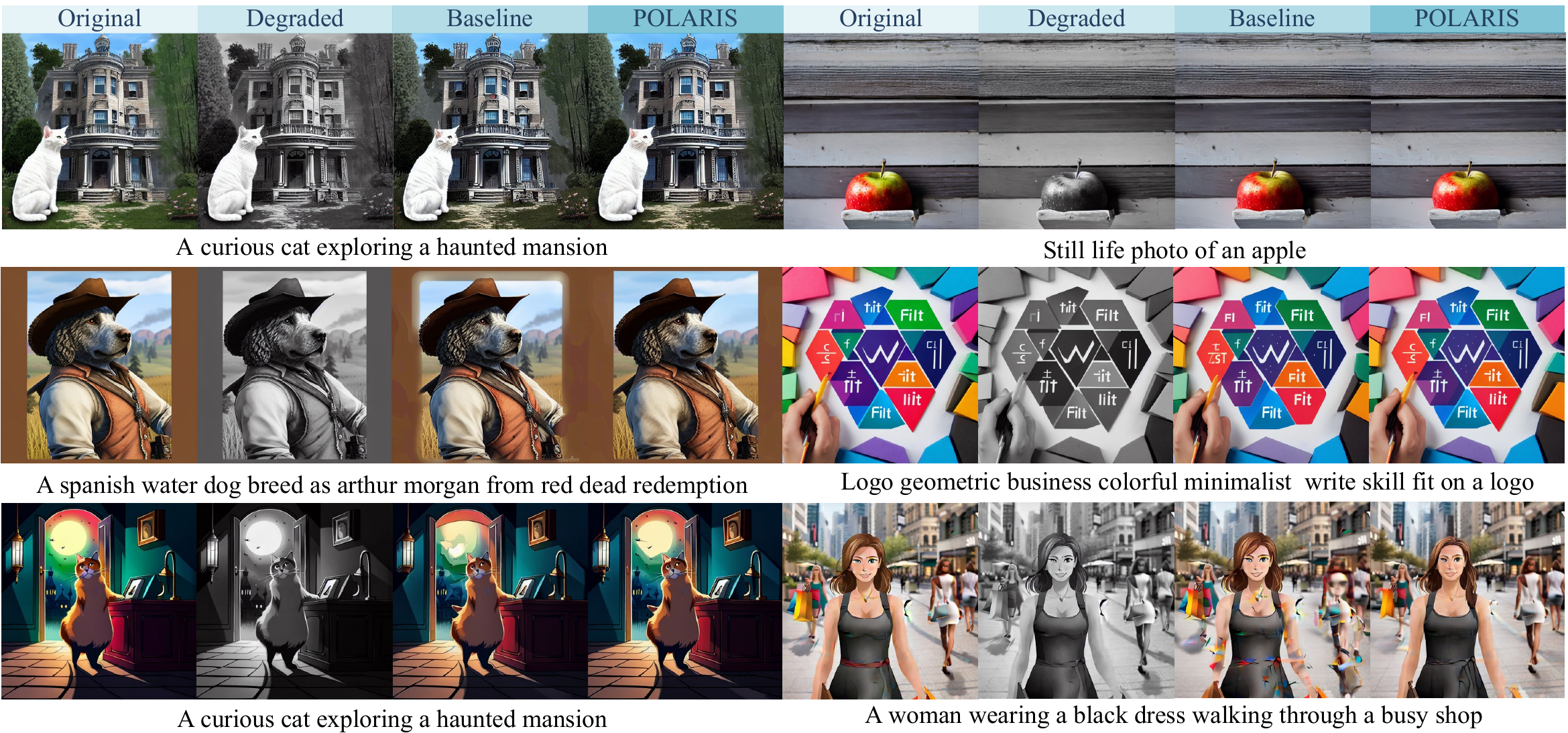}
    \caption{\textbf{Image Colorization} visualizations}
    \label{fig:supp_color}
\end{figure}

Figs.~\ref{fig:editing_viz}–\ref{fig:editing_viz2} present additional qualitative results on text-guided image editing, showing that our method preserves background details while performing complex semantic modifications. Figs.~\ref{fig:supp_deblurring}–\ref{fig:supp_color} further demonstrate its effectiveness across deblurring, super-resolution, inpainting, and colorization tasks. Note: \textbf{The “Degraded’’ columns contain decoded initial latents, rather than pixel-space degradations, to reveal the true starting point of the reverse diffusion and the challenges inherent to the latent space.}

\vspace{1em}

\vspace{1em}

\newpage